\documentclass[preprint,12pt]{elsarticle}

\journal{Physics Letters A}

\usepackage[numbers]{natbib}
\usepackage{microtype}
\usepackage{graphicx}
\usepackage{subcaption}
\usepackage{todonotes}
\usepackage{booktabs} 
\usepackage{bm}
\usepackage{bbm}
\usepackage{amsthm}

\usepackage{hyperref}

\newcommand{\npb}[1]{#1}
\newcommand{\physica}[1]{\textcolor{black}{#1}}

\renewcommand{\vec}[1]{\bm{#1}}

\usepackage{amsmath,amsfonts,bm}

\def\eqref#1{equation~\ref{#1}}

\def\1{\bm{1}}

\def\vw{{\bm{w}}}

\def\vphi{{\bm{\phi}}}

\def\mH{{\bm{H}}}

\DeclareMathAlphabet{\mathsfit}{\encodingdefault}{\sfdefault}{m}{sl}
\SetMathAlphabet{\mathsfit}{bold}{\encodingdefault}{\sfdefault}{bx}{n}

\begin{document}

\begin{frontmatter}



\title{Appearance of random matrix theory in deep learning}


\author[bris]{Nicholas P. Baskerville}

\affiliation[bris]{organization={School of Mathematics, University of Bristol},
            addressline={Fry Building, Woodland Road}, 
            city={Bristol},
            postcode={BS8 1UG}, 
            country={United Kingdom}}

\author[oxml]{Diego Granziol}
\author[oxmi]{Jonathan P. Keating}

\affiliation[oxml]{organization={Machine Learning Research Group, University of Oxford},
            addressline={Walton Well Road}, 
            city={Oxford},
            postcode={OX2 6ED}, ,
            country={United Kingdom}}
            
\affiliation[oxmi]{organization={Mathematical Institute, University of Oxford},
            addressline={Andrew Wiles Building, Woodstock Road}, 
            city={Oxford},
            postcode={OX2 6GG}, ,
            country={United Kingdom}}

\begin{abstract}
We investigate the local spectral statistics of the loss surface Hessians of artificial neural networks, where we discover agreement with Gaussian Orthogonal Ensemble statistics across several network architectures and datasets. These results shed new light on the applicability of Random Matrix Theory to modelling neural networks and suggest a role for it in the study of loss surfaces in deep learning. 
\end{abstract}
\end{frontmatter}

\section{Introduction}
Artificial Neural Networks (ANNs) continually advance the state of the art in machine learning, including computer vision, speech processing and natural language processing. However, we do not have a precise theoretical understanding of their training and generalisation dynamics. The observation that gradient based optimisation methods \cite{Bottou2012} with different random initialisations do not seem to get stuck in poor quality local minima, despite the high dimensionality and non-convexity of the optimisation problems, has led to a significant focus on neural network \emph{loss surfaces}. 

\npb{The loss of a neural network is a scalar function that measures how well the network is performing on a particular data item, with lower values being better. Loss functions are defined to be greater than or equal to zero. For example, a neural network's output may be the predicted house price given a set of features about the house, and the loss value for a particular house could be the squared error of the predicted price compared to the known true price. The loss surface of a neural network is the value of the loss, averaged over the data (houses in the above example) and viewed as a function of the network \emph{weights}, i.e. its free parameters. The task of optimisation for neural networks amounts to finding low-points of the loss surface in a weight space which is typically very high dimensional (e.g. $10^7$ is not uncommon in practice). \physica{The loss surface in most realistic examples will be non-convex and possess many local minima and saddle points (though one notable exception is the common practice of training only the final layer of a deep network). The loss is almost always optimised using stochastic gradient descent (or some variant thereof such as Adam \citep{kingma2014adam} or Adagrad \citep{duchi2011adaptive}, which use a per-parameter learning rate which depends on the running covariance of the gradients) and so} one expects the local minima and low index saddle points to be important, being the points where the optimisation is likely to become trapped.}

The loss surface is typically investigated through the Hessian \physica{which is the second order taylor expansion of the loss and hence especially relevant at local minima}. Under strong simplifying assumptions, such as independence of the neural network inputs and weights \citep{choromanska2015loss,choromanska2015open,pennington2017geometry}, the Hessian at critical points  of the loss (where the gradient is zero), is described by certain important classes of random matrices, such as the Gaussian Orthogonal Ensemble (\emph{GOE}) \citep{tao2012topics} or the Wishart Ensemble \citep{bun2017cleaning} of Random Matrix Theory (RMT). The average spectral density (taken over an ensemble) of these matrices, in the limit of infinite dimension, can be calculated; for the GOE the result is known as the {\em Wigner semicircle law}, and for the Wishart Ensemble it is the {\em Marchenko-Pastur law}.  Hence with these assumptions, one can make quantitative predictions about the nature of the critical points and aspects of the geometry of the loss landscape. 

\npb{Several authors have studied the similarity between types of neural networks and models from statistical physics, such as spin glasses starting with \citet{amit1985spin, gardner1988optimal}, and more recently connections with the ANNs of machine learning, both empirical \citep{sagun2014explorations} and theoretical \citep{choromanska2015loss}.  \citet{chaudhari2015energy} introduce connections with magnetic fields in disordered systems, demonstrating an analogy with weight decay in DNNs.} \physica{Connections between machine learning and statistical physics from various viewpoints are detailed extensively in \cite{bahri2020statistical, mezard2009information, zdeborova2016statistical, carleo2019machine, gabrie2020mean, roberts2021principles}.}

\begin{figure*}[t!]
    \centering
    \begin{subfigure}{0.32\linewidth}
        \includegraphics[width=1\linewidth,trim={0 0 0 0},clip]{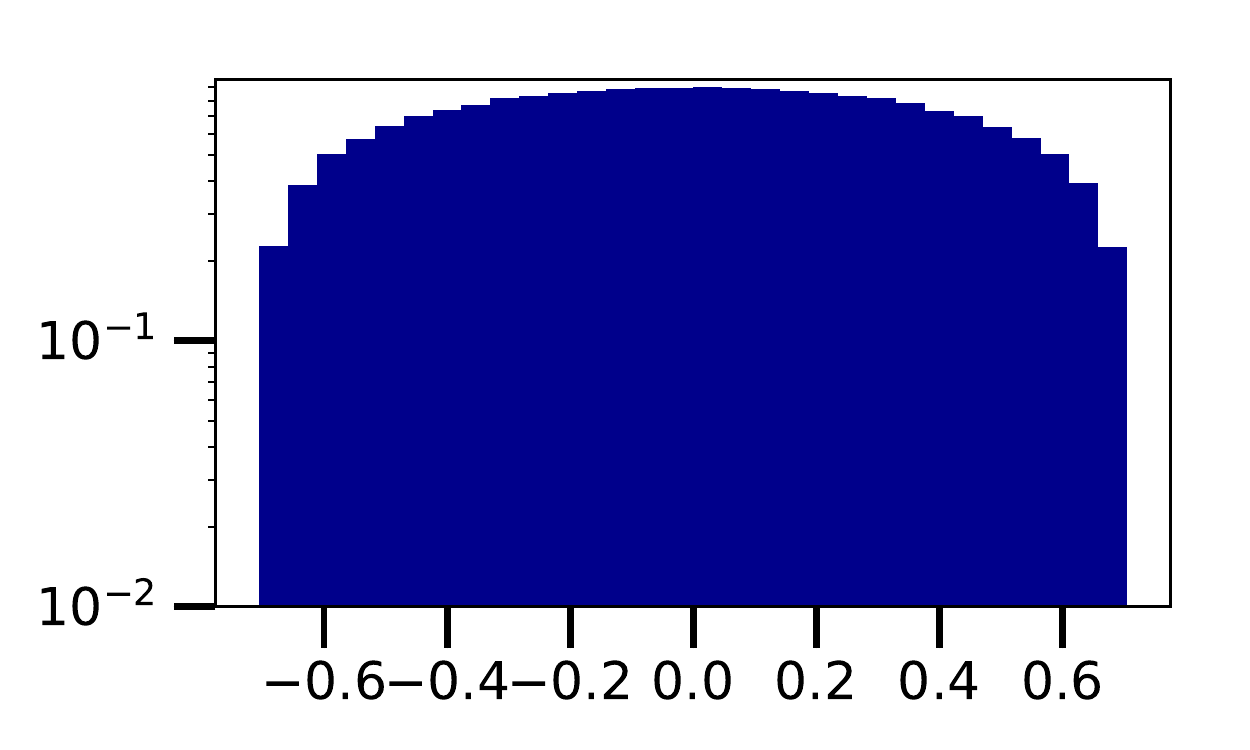}
        \caption{Wigner semicircle}
        \label{subfig:wignerstemintro}
    \end{subfigure}
    \begin{subfigure}{0.32\linewidth}
        \includegraphics[width=1\linewidth,trim={0 0 0 0},clip]{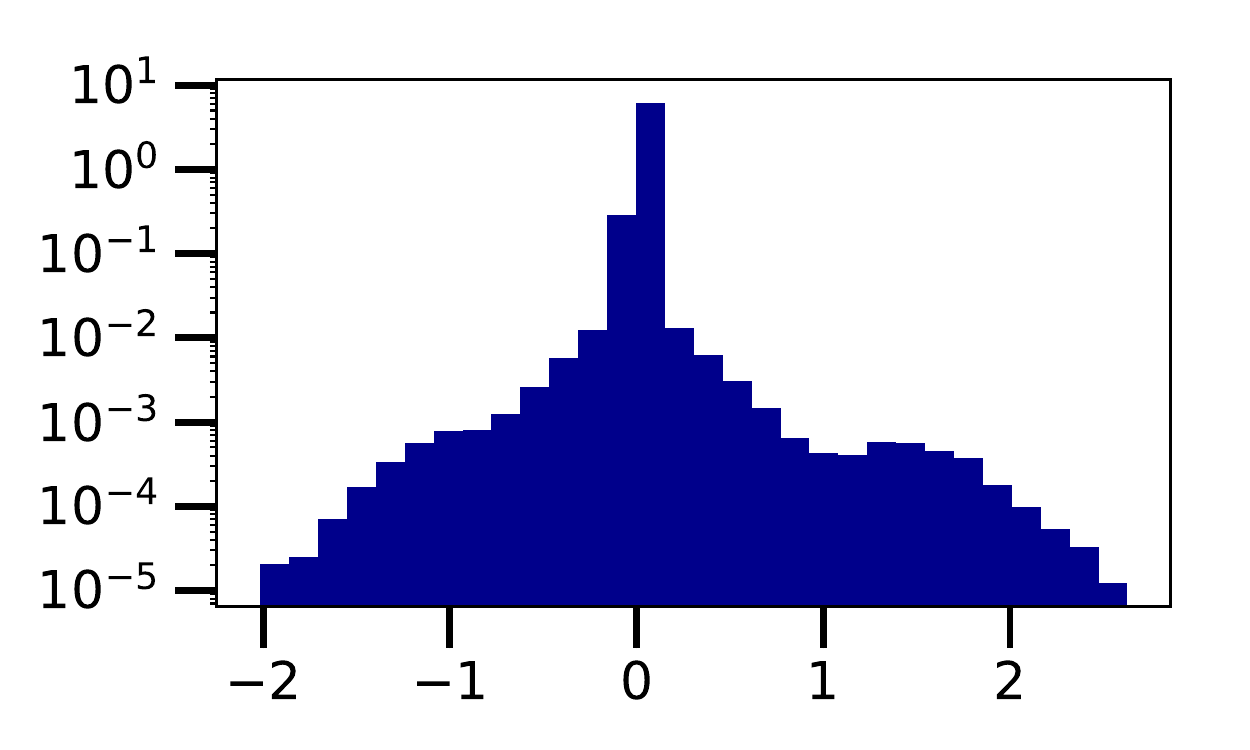}
        \caption{MLP}
        \label{subfig:mlp}
    \end{subfigure}
    \begin{subfigure}{0.32\linewidth}
        \includegraphics[width=1\linewidth,trim={0 0 0 0},clip]{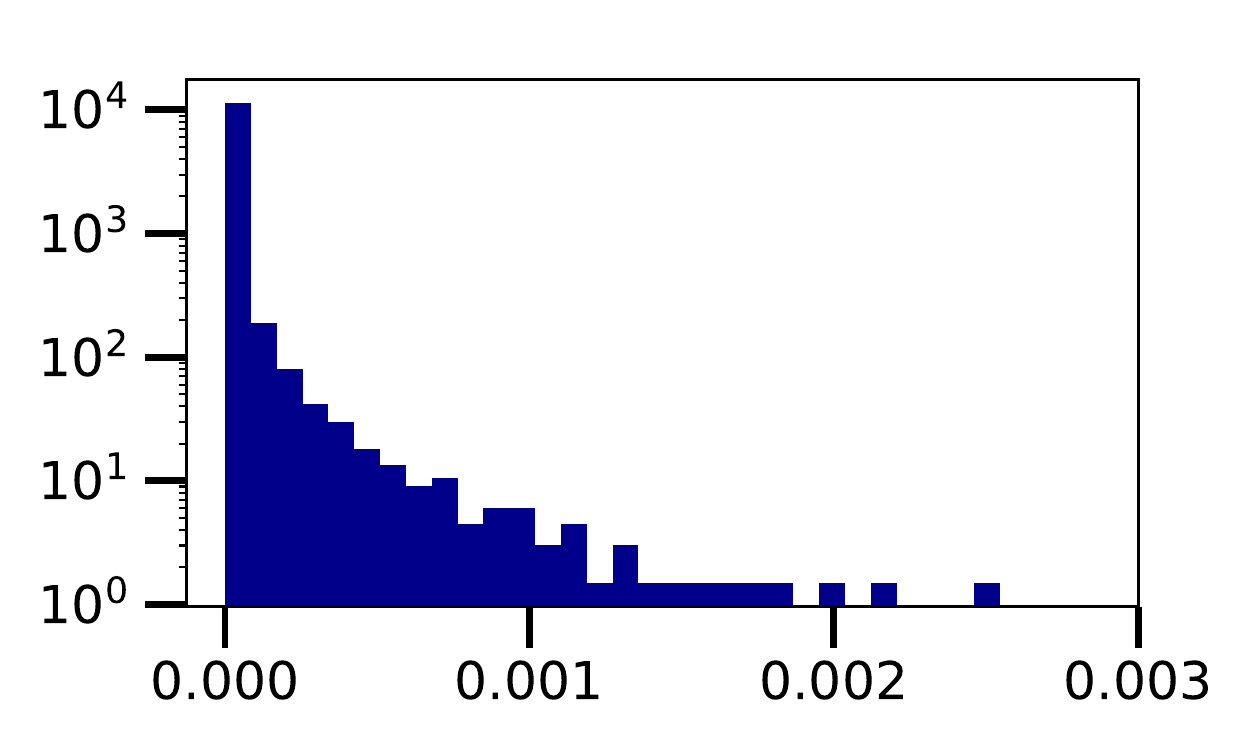}
        \caption{Logistic Regression}
        \label{subfig:logisticregression}
    \end{subfigure}
    \caption{\physica{Comparison of different global spectral statistics (spectral densities). (a) We show actual GOE data to demonstrate the form of the Wigner semicircle. (b) Hessian of cross entropy loss for MLP on MNIST. (c) Hessian of cross entropy loss for logistic regression on MNIST.} Note the log-scale on the y-axis. A few outliers have been clipped from logistic regression to aid visualisation.}\label{fig:goeisabadfit}
\end{figure*}

\citet{choromanska2015loss} showed, assuming i.i.d Gaussian inputs and network path independence, that a multi-layer ReLU neural network's loss is equivalent to that of a spin-glass model. Its conditional Hessian spectrum is thus given by a GOE calculation \citep{auffinger2013random} involving real-symmetric matrices with otherwise independent Gaussian random entries. It follows that under these assumptions local minima are located within a narrow band, bounded below by the global minimum. The practical implication is that for a sufficient number of hidden layers (more than 2) all local minima are \textit{close} in loss to the global minimum. \citet{baskerville2021loss} extend this line of work to networks with general activation functions. \citet{baskerville2021spin} show for Generative Adversarial Networks, using a spin-glass model for both the generator and discriminator, that the structure of local optima encourages collapse to a narrow band of the loss for at least one of the networks but not necessarily both simultaneously. \npb{These works are examples of complexity calculations, which have a considerable history in the physics and mathematics literature \citep{fyodorov2004complexity, fyodorov2007replica, fyodorov2014topology}. Recent works have completed complexity calculations and studied properties of local minima of various models intended to highlight aspects of loss surfaces in high dimensions \citep{ros2019complex, NEURIPS2019_fbad540b, maillard2020landscape,fyodorov2018hessian}.}
Similarly, \citet{pennington2017geometry} use the Gauss Newton decomposition of a squared loss Hessian, assuming independence and normality of both the data and weights along with free addition of the resulting Wigner/Wishart ensembles, to derive a functional form for the critical index (the fraction of the eigenvalues that are negative) as a function of the loss. They show that below a certain critical energy threshold \textit{all critical points are minima}. 

\physica{Several works have used randomised models of neural networks to study properties of the training and test loss, such as the double-descent phenomenon\footnote{Increasing the network size initially leads to over-fitting, but beyond a critical point further increasing the network size decreases the test error to a lower level than the optimal small network.}. The phenomenon can be recovered using randomised models of neural networks, such as random feature models (two-layer networks training only the last layer where the output of the first layer is i.i.d. Gaussian) \cite{mei2019generalization, gerace2020generalisation, dhifallah2020precise}. \citet{ba2020generalization} extended this analysis to two layer networks where either of the layers can be trained and demonstrated that the double-descent cannot be recovered when training only the lower layer. Combined with the work of \citet{adlam2020neural}, the emerging understanding is that important properties of neural networks observed in practice can be recreated using simplified randomised models and random matrix theory, but the picture is far from complete.}

\npb{Pennington and Worah have studied Gram matrices of network outputs \citep{pennington2019nonlinear} and also neural network Fisher information matrices \citep{FisherRMT} in the context of single hidden layer networks with i.i.d. Gaussian weights and inputs. \citet{benigni2019eigenvalue} extended this work to any number of layers and i.i.d. weights and inputs with sub-Gaussian tails.}

An important and fundamental problem with the aforementioned works is that typically the average spectral density of the Hessian of neural networks does not in fact match that of the associated random matrix ensembles.  This is illustrated in Figure \ref{fig:goeisabadfit}. 
Put simply, \emph{one does not observe the Wigner semicircle or Marchenko-Pastur eigenvalue distributions, implied by the Gaussian Orthogonal or Wishart Ensembles for ANNs}. 
As shown in \citet{granziol2020beyond,granziol2019towards,papyan2018full,papyan2019measurements,ghorbani2019investigation,sagun2016eigenvalues,sagun2017empirical} the spectral density of ANN Hessians contain outliers and a large number of near zero eigenvalues, features not seen in canonical random matrix ensembles. 
Furthermore, even allowing for this, as shown in \cite{granziol2020towards} by specifically embedding outliers as a low rank perturbation to a random matrix, the remaining bulk spectral density still does not match the Wigner semicircle or Marchenko-Pastur distributions \cite{granziol2020beyond}, bringing into question the validity of the underlying modelling.  

The fact that the experimental results differ markedly from the theoretical predictions has called into question the validity of ANN analyses based on canonical random matrix ensembles. Moreover, the compelling results of works such as \cite{choromanska2015loss, pennington2017geometry} are obtained using very particular properties of the canonical ensembles, such as large deviation principles, as pointed out in \citet{granziol2020beyond}. The extent to which such results can be generalised is an open question.
Hence, further work is required to better understand to what extent Random Matrix Theory can be used to analyse the loss surfaces of ANNs.


In the present paper, we show that the  {\em local spectral statistics} (i.e.~those measuring correlations on the scale of the mean eigenvalue spacing) of ANN Hessians are well modelled by those of  GOE random matrices, even when the mean spectral density is different from the semicicle law. We display these results experimentally on MNIST trained multi-layer perceptrons and on the final layer of a ResNet-$34$ on CIFAR-$10$. \npb{The objective of our work is to motivate a new use for Random Matrix Theory in the study of the theory of deep neural networks.}.  In the context of more established applications of Random Matrix Theory, this conclusion may not be so surprising -- it has often been observed that the local spectral statistics are universal while the mean density is not -- however, in the context of Machine Learning this important point has not previously been made, nor its consequences explored. Our main goal is to illustrate it in that setting, through numerical experiments, and to start to examine some of its implications.

\section{Preliminaries}
\label{sec:prelim}
Consider a neural network with weights $\vec{w}\in\mathbb{R}^P$ and a dataset with distribution $\mathbb{P}_{\mathrm{data}}$. \npb{For the purposes of our discussion, a neural network, $f_{\vec{w}}$ say, is just a non-linear function from some $\mathbb{R}^d$ to some $\mathbb{R}^c$, parametrised by $\vec{w}$. Neural networks can be defined in many different ways in terms of their weights (the architecture of the network), but these details will not play role in our discussion. What will be important is that the number of weights $P$ will be large, i.e. approaching 10,000 even in the simplest of cases.} Let $L(\vec{w}, \vec{x})$ be the loss of the network for a single datum $\vec{x}$ and let $\mathcal{D}$ denote any finite sample of data points from $\mathbb{P}_{\mathrm{data}}$. \npb{A simple example of $L$ is the squared error $L(\vec{w}, (\vec{x}, \vec{y})) = ||f_{\vec{w}}(\vec{x}) - \vec{y}||_2^2$, where $\mathbb{P}_{\mathrm{data}}$ is a distribution on tuples of features $\vec{x}$ and labels $\vec{y}$.} The \emph{true loss} is given by \begin{align}
    \mathcal{L}_{true}(\vec{w}) = \mathbb{E}_{\vec{x}\sim\mathbb{P}_{\mathrm{data}}} L(\vec{w}, \vec{x})
\end{align}
and the \emph{empirical loss} (or training loss) is given by \begin{align}
        \mathcal{L}_{emp}(\vec{w}, \mathcal{D}) = \frac{1}{|\mathcal{D}|}\sum_{\vec{x}\in\mathcal{D}} L(\vec{w}, \vec{x}).
\end{align}
Where $\mathcal{D}$ denotes the dataset. The true loss is a deterministic function of the weights, while the empirical loss is a random function with the randomness coming from the random sampling of the finite dataset $\mathcal{D}$. The empirical Hessian $\mH_{emp}(\vw) = \nabla^{2} \mathcal{L}_{emp}(\vw)$, describes the loss curvature at the point $\vw$ in weight space. By the spectral theorem, the Hessian can be written in terms of its eigenvalue/eigenvector pairs $\mH_{emp} = \sum_{i}^{P}\lambda_{i}\vphi_{i}\vphi_{i}^{T}$, where the dependence on $\vw$ has been dropped to keep the notation simple. The eigenvalues of the Hessian are particularly important, being explicitly required in second-order optimisation methods, and characterising the stationary points of the loss as local minima, local maxima or generally saddle points of some other index.

For a matrix drawn from a probability distribution, its eigenvalues are random variables. The eigenvalue distribution is described by the joint probability density function (j.p.d.f) $p(\lambda_1, \lambda_2, \ldots, \lambda_P)$, also known as the $P$-point correlation function.
The simplest example is the \emph{empirical spectral density (ESD)}, $\rho^{(P)}(\lambda) = \frac{1}{P}\sum_{i}^{P}\delta(\lambda-\lambda_{i})$.  Integrating $\rho^{(P)}(\lambda)$ over an interval with respect to $\lambda$ gives the fraction of the eigenvalues in that interval.   
Taking an expectation over the random matrix ensemble, we obtain the \emph{mean spectral density} $\mathbb{E}\rho^{(P)}(\lambda)$, which is a deterministic probability distribution on $\mathbb{R}$. Alternatively, taking the $P\rightarrow\infty$ limit, assuming it exists, gives the \emph{limiting spectral density (LSD)} $\rho$, another deterministic probability distribution on $\mathbb{R}$. A key feature of many random matrix ensembles is \emph{self-averaging} or \emph{ergodicity}, meaning that the leading order term (for large $P$) in $\mathbb{E}\rho^{(P)}$ agrees with $\rho$. Given the j.p.d.f, one can obtain the mean spectral density, known as the $1$-point correlation function (or any other $k$-point correlation function) by marginalisation \begin{equation}
    \mathbb{E}\rho^{(P)}(\lambda) = \int p(\lambda, \lambda_2, \ldots, \lambda_P) d\lambda_2\ldots d\lambda_P.
\end{equation}A GOE matrix is an example of a \emph{Wigner random matrix}, namely a real-symmetric (or complex-Hermitian) matrix with otherwise i.i.d. entries and off-diagonal variance $\sigma^2$.\footnote{The GOE corresponds to taking the independent matrix entries to be normal random variables.} The mean spectral density for Wigner matrices is known to be Wigner's semicircle \cite{mehta2004random} \begin{equation}
    \rho_{SC}(\lambda) = \frac{1}{2\pi \sigma^2 P}\sqrtsign{4P\sigma^2 - \lambda^2}\mathbf{1}_{|\lambda|\leq 2\sigma\sqrtsign{P}}.
\end{equation} The radius of the semicircle\footnote{Using the Frobenius norm identity $\sum_{i}^{P}\lambda_{i}^{2} = P^{2}\sigma^{2}$} is proportional to $\sqrtsign{P}\sigma$, hence scaling Wigner matrices by $1/\sqrt{P}$ leads to a limit distribution when $P\rightarrow\infty$.  This is the LSD. 
With this scaling, there are, on average, $\mathcal{O}(P)$ eigenvalues in any open subset of the compact spectral support. In this sense, the mean (or limiting) spectral density is \emph{macroscopic}, meaning that, as $P\rightarrow\infty$, one ceases to see individual eigenvalues, but rather a continuum with some given density.

\section{Motivation: Microscopic Universality}

Random Matrix Theory was first developed in physics to explain the statistical properties of nuclear energy levels, and later used to describe the spectral statistics in atomic spectra, condensed matter systems, quantum chaotic systems etc; see, for example \citep{weidenmuller2008random, beenakker1997random, berry1987quantum, bohigas1991random}. \emph{None of these physical systems exhibits a semicircular empirical spectral density}. However they all generically show agreement with RMT at the level of the mean eigenvalue spacing when local spectral statistics are compared.  Our point is that while neither multi-layer perceptron (MLP) nor Softmax Regression Hessians are described by the Wigner semicircle law which holds for GOE matrices (c.f. Figure 1a) -- their spectra contain outliers, large peaks near the origin and the remaining components of the histogram also do not match the semicircle -- \physica{nevertheless Random Matrix Theory can still (and we shall demonstrate does) describe spectral fluctuations on the scale of their mean eigenvalue spacing}.

\medskip
\physica{It is worth noting in passing that possibilities other than random-matrix statistics exist and occur. For example, in systems that are classically integrable, one finds instead Poisson statistics \cite{berry1977level, berry1987quantum}; similarly, Poisson statistics also occur in disordered systems in the regime of strong Anderson localisation \cite{efetov1999supersymmetry}; and for systems close to integrable one finds a superposition of random-matrix and Poisson statistics \cite{berry1984semiclassical}. So showing that Random Matrix Theory applies is far from being a trivial observation.  Indeed it remains one of the outstanding challenges of mathematical physics to prove that the spectral statistics of any individual Hamiltonian system are described by it in the semiclassical limit.}

\medskip
Physics RMT calculations re-scale the eigenvalues to have a mean level spacing of $1$ and then typically look at the \emph{nearest neighbour spacings distribution} (NNSD), i.e. the distribution of the distances between adjacent pairs of eigenvalues.  One theoretical motivation for considering the NNSD is that it is independent of the Gaussianity assumption and reflects the symmetry of the underlying system. It is the NNSD that is universal (for systems of the same symmetry class) and not the average spectral density, which is best viewed as a parameter of the system. The aforementioned transformation to give mean spacing $1$ is done precisely to remove the effect of the average spectral density on the pair correlations leaving behind only the universal correlations. To the best of our knowledge no prior work has evaluated the NNSD of artificial neural networks and this is a central focus of this paper.

\begin{figure}[h]
    \centering
    \includegraphics[width=0.8\textwidth]{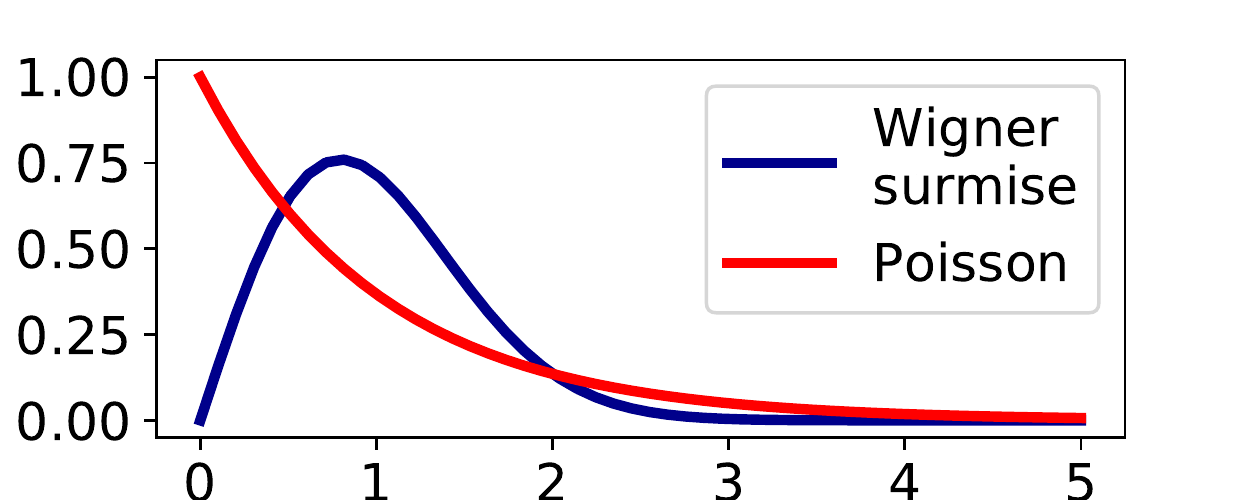}
    \caption{The density of the Wigner surmise.}
    \label{fig:wigner}
\end{figure}

In contrast to the LSD, other $k$-point correlation functions are also normalised such that the mean spacing between adjacent eigenvalues is unity. At this \emph{microscopic} scale, the LSD is locally constant and equal to 1 meaning that its effect on the eigenvalues' distribution has been removed and only microscopic correlations remain. In the case of Wigner random matrices, for which the LSD varies slowly across the support of the eigenvalue distribution, this corresponds to scaling by $\sqrt{P}$. On this scale the limiting eigenvalue correlations when $P\to\infty$ are {\em universal}; that is, they are the same for wide classes of random matrices, depending only on symmetry \cite{guhr1998random}.  For example, this universality is exhibited by the NNSD. Consider a $2\times 2$ GOE matrix, in which case the j.p.d.f has a simple form: \begin{equation}
    p(\lambda_1, \lambda_2) \propto |\lambda_1 - \lambda_2| e^{-\frac{1}{2}(\lambda_1^2 + \lambda_2^2)}.
\end{equation}
Making the change of variables $\nu_1 = \lambda_1 - \lambda_2, \nu_2 = \lambda_1 + \lambda_2$, integrating out $\nu_2$ and setting $s = |\nu_1|$ results in a density $\rho_{Wigner}(s) = \frac{\pi s}{2}e^{-\frac{\pi}{4}s^2}$, known as the \emph{Wigner surmise} (see Figure \ref{fig:wigner}). For larger matrices, the j.p.d.f must include an indicator function $\mathbbm{1}\{\lambda_1\leq \lambda_2\leq \ldots \lambda_P\}$ before marginalisation so that one is studying pairs of \emph{adjacent} eigenvalues. While the Wigner surmise can only be proved exactly, as above, for the 2 × 2 GOE, it holds to high accuracy for the NNSD of GOE matrices of any size provided that the eigenvalues have been scaled to give mean spacing 1.\footnote{An exact formula for the NNSD of GOE matrices of any size, and one that holds in the large $P$ limit, can be found in \citet{mehta2004random}.} The Wigner surmise density vanishes at $0$, capturing `repulsion' between eigenvalues that is characteristic of RMT statistics, in contrast to the distribution of entirely independent eigenvalues given by the \emph{Poisson law} $\rho_{Poisson}(s) = e^{-s}$. The Wigner surmise is universal in that the same density formula applies to all real-symmetric random matrices, not just the GOE or Wigner random matrices. 

\section{Methodology}\label{sec:methods}
Prior work \citep{granziol2019mlrg,papyan2018full,ghorbani2019investigation} focusing on the Hessian empirical spectral density has utilised fast Hessian vector products \cite{pearlmutter1994fast} in conjunction with Lanczos \cite{meurant2006lanczos} methods. 
However, these methods approximate only macroscopic quantities like the spectral density, not microscopic statistics such as nearest neighbour spectral spacings. 
For modern neural networks, the $\mathcal{O}(P^{3})$ Hessian eigendecomposition cost will be prohibitive,  e.g. for a Residual Network \npb{(Resnet) \cite{he2016deep}} with $34$ layers $P=10^{7}$. Hence, We restrict to models small enough to perform exact full Hessian computation and eigendecomposition. 

\medskip
We consider single layer neural networks for classification (softmax regression), 2-hidden-layer MLPs\footnote{Hidden layer widths: 10, 100.} and 3 hidden-layer MLPs\footnote{Hidden layer widths: 10, 100, 100.}. On MNIST \citep{deng2012mnist}, the Hessians are of size $7850\times 7850$ for logistic regression, $9860\times 9860$ for the small MLP and $20060\times 20060$ for the larger 3 hidden-layer MLP, so can be computed exactly by simply applying automatic differentiation twice, and the eigenvalues can be computed exactly in a reasonable amount of time.
We also consider a single layer applied to CIFAR-$10$  \citep{krizhevsky2009learning} classification with pre-trained Resnet-$34$ embedding features \npb{\cite{he2016deep, resnet-torch}}. While we cannot at present study the full Hessian of, for example, a Resnet-$34$, we can study the common transfer learning use-case of training only the final layer on some particular task \cite{sharif2014cnn}. The Hessians can be computed at any data point or over any collection of data points. We consider Hessians computed over the entire datasets in question, and over batches of size 64. We separately consider test and train sets.

\medskip
\physica{In order to extend the relevance of our analysis to beyond logistic regression and MLP, we consider one of the simplest convolutional neural networks (CNN) of the form of LeNet \citep{lecun1998mnist} on CIFAR-$10$. Compared to the standard LeNet (which has over $50000$ parameters) we reduce the number of neurons in the first fully connected layer from $120$ to $35$ and the second from $84$ to $50$. Note that the resulting architecture contains a bottleneck in the intermediate layer, in contrast to the ``hour-glass'' shapes that are necessary to maintain manageable parameter numbers with full MLP architectures. Despite reducing the total number of parameters by a factor of $3$ we find the total validation accuracy drop to be no more than $2\%$. The total validation accuracy of $69\%$ is significantly below state of the art $\approx 95\%$, but we are clearly in the regime where significant learning can and does take place, which we consider sufficient for the purposes of this manuscript. We also extend our experiments beyond the cross entropy loss function, by considering a regression problem ($L_{2}$ loss) and beyond the high-dimensional feature setting of computer vision with the Bike dataset\footnote{\href{https://archive.ics.uci.edu/ml/datasets/Bike+Sharing+Dataset}{https://archive.ics.uci.edu/ml/datasets/Bike+Sharing+Dataset} (accessed 14/10/21)} which has only 13-dimensional feature vectors and a single-dimensional regressand (see Appendix \ref{app:preproc} for details of our data pre-processing). The architecture in this case widens considerably in the first layer (from 13 inputs to 100 neurons) and that gradually tapers to the single output. The final test loss (i.e. mean squared error) of the trained model is 0.044 which is competitive with baseline results \cite{wang2019exact}\footnote{\citet{wang2019exact} report an RMSE of 0.220 on Bike (which corresponds to 0.048 mean squared error) using a Gaussian process regression model with exact inference.}}

\paragraph{Training details:}All networks were trained using SGD for 300 epochs with initial learning rate 0.003, linear learning rate decay to 0.00003 between epoch 150 and 270, momentum 0.9 and weight decay $5\times 10^{-4}$. We use a PyTorch \citep{paszke2017automatic} implementation. Full code to reproduce our results is made available \footnote{\url{https://github.com/npbaskerville/dnn-rmt-spacings}}. Full descriptions of all network architectures are given in the Appendix \ref{app:architectures}.

\section{Spectral spacing statistics in RMT}
\label{sec:spacings}
Consider a random $P\times P$ matrix $M_P$ with ordered $\lambda_1 \leq \lambda_2 \leq \ldots \leq \lambda_P$. Let $I_{ave}$ be the mean spectral cumulative density function for the random matrix ensemble from which $M_P$ is drawn. The \emph{unfolded spectrum} is defined as \begin{align}
    l_i = I_{ave}(\lambda_i).
\end{align}
The unfolded spacings are then defined as \begin{align}
    s_i = l_i - l_{i-1}, ~~~ i=2, \ldots, P.
\end{align}

With this definition, the mean of the $s_i$ is unity, which means that this transformation has brought the eigenvalues on to the microscopic scale on which universal spectral spacing statistics emerge. We are investigating the presence of Random Matrix Theory statistics in neural networks by considering the nearest neighbour spectral spacings of their Hessians. Within the Random Matrix Theory literature, it has been repeatedly observed \citep{bohigas1991random, berry1987quantum} that the unfolded spacings of a matrix with RMT pair correlations follow universal distributions determined only by the symmetry class of the $M_P$. Hessians are real symmetric, so the relevant universality class is GOE and therefore the unfolded neural network spacings should be compared to the Wigner surmise
\begin{equation}\label{eq:wigner}
    \rho_{Wigner}(s) = \frac{\pi s}{2}e^{-\frac{\pi}{4}s^2}.
\end{equation}
A collection of unfolded spacings $s_2,\ldots, s_P$ from a matrix with GOE spacing statistics should look like a sample of i.i.d. draws from the Wigner surmise density (\ref{eq:wigner}).
For some known random matrix distributions, $I_{ave}$ may be available explicitly, or at least via highly accurate quadrature methods from a known mean spectral density. For example, for the $P\times P$ GOE  \citep{abuelenin2012effect} $I_{ave}^{GOE}(\lambda) $ is given by:
\begin{align}
   P\left[\frac{1}{2} + \frac{\lambda}{2\pi P}\sqrt{2P - \lambda^2} + \frac{1}{\pi}\arctan\left(\frac{\lambda}{\sqrt{2 P - \lambda^2}}\right)\right].
\end{align}
However, when dealing with experimental data where the mean spectral density is unknown, one must resort to using an approximation to $I_{ave}$. Various approaches are used in the literature, including polynomial spline interpolation \citep{abuelenin2012effect}. The approach of \citep{scholak2014spectral, unfoldr} is most appropriate in our case, since computing Hessians over many mini-batches of data results in a large pool of spectra which can be used to accurately approximate $I_{ave}$ simply by the empirical cumulative density.
Suppose that we have $m$ samples $(M^{(i)}_P)_{i=1}^m$ from a random matrix distribution over symmetric $P\times P$ matrices. Fix some integers $m_1, m_2 > 0$ such that $m_1 + m_2 = m$. The spectra of the matrices $(M^{(i)}_P)_{i=1}^{m_1}$ can then be used to construct an approximation to $I_{ave}$.  More precisely, let $\Lambda_1$ be the set of all eigenvalues of the $(M^{(i)}_P)_{i=1}^{m_1}$, then we define \begin{align}
    \tilde{I}_{ave}(\lambda) = \frac{1}{|\Lambda_1|} |\{\lambda' \in \Lambda_1 \mid \lambda' < \lambda\}|.
\end{align}
For each of the matrices  $(M^{(i)}_P)_{i=m_1 + 1}^m$, one can then use $\tilde{I}_{ave}$ to construct their unfolded spacings. When the matrix size $P$ is small, one can only study the spectral spacing distribution by looking over multiple matrix samples. However, the same spacing distribution is also present for a single matrix in the large $P$ limit. A clear disadvantage of studying unfolded nearest neighbour spectral spacings with the above methods is the need for a reasonably large number of independent matrix samples. This rules-out studying the unfolded spacings of a single large matrix. Another obvious disadvantage is the introduction of error by the approximation of $I_{ave}$, giving the opportunity for local spectral statistics to be distorted or destroyed. An alternative statistic is the consecutive spacing ratio of \citep{atas2013distribution}. In the above notation, the ratios for a single $P\times P$ matrix are defined as \begin{align}
   r_i= \frac{\lambda_{i} - \lambda_{i-1}}{\lambda_{i-1} - \lambda_{i-2}}, ~~ 2 \leq i \leq P.
\end{align}
\citet{atas2013distribution} proved a `Wigner-like surmise' for the spacing ratios, which for the GOE is \begin{align}
P(r) = \frac{27(r + r^2)}{8(1 + r + r^2)^{5/2}}.
\end{align}

In our experiments, we can compute the spacing ratios for Hessians computed over entire datasets or over batches, whereas the unfolded spacing ratios can only be computed in the batch setting, in which case \physica{a random} $\frac{2}{3}$ of the batch Hessians are reserved for computing $\tilde{I}_{ave}$ and the remaining $\frac{1}{3}$ are unfolded and analysed. \physica{This split is essentially arbitrary, except that  we err on the side of using more to compute $\tilde{I}_{ave}$ since even a single properly unfolded spectrum can demonstrate universal local statistics.}

\section{Results}
We display results as histograms of data along with a plot of the Wigner (or the Wigner-like) surmise density. We make a few practical adjustments to the plots.  Spacing ratios are truncated above some value, as the presence of a few extreme outliers makes visualisation difficult. We choose a cut-off at 10. Note that around 0.985 of the mass of the Wigner-like surmise is below 10, so this is a reasonable adjustment. The hessians have degenerate spectra. The Wigner surmise is not a good fit to the observed unfolded spectra if the zero eigenvalues are retained. Imposing a lower cut-off of $10^{-20}$ in magnitude is sufficient to obtain agreement with Wigner.\footnote{\physica{For example, in the case of the 3-hidden-layer MLP on MNIST shown in Figure \ref{fig:deep_mlp_spacings}, among 157 batch-wise spectra the proportion of eigenvalues below the cut-off was between $0.29$ and $0.40$.}} This is below the machine precision, so these omitted eigenvalues are indistinguishable from 0.
\begin{figure}[h!]
\centering
\begin{tabular}{cc}
\subfloat[Unfolded spacings. Batch-size 64.]{\includegraphics[width=0.4\textwidth]{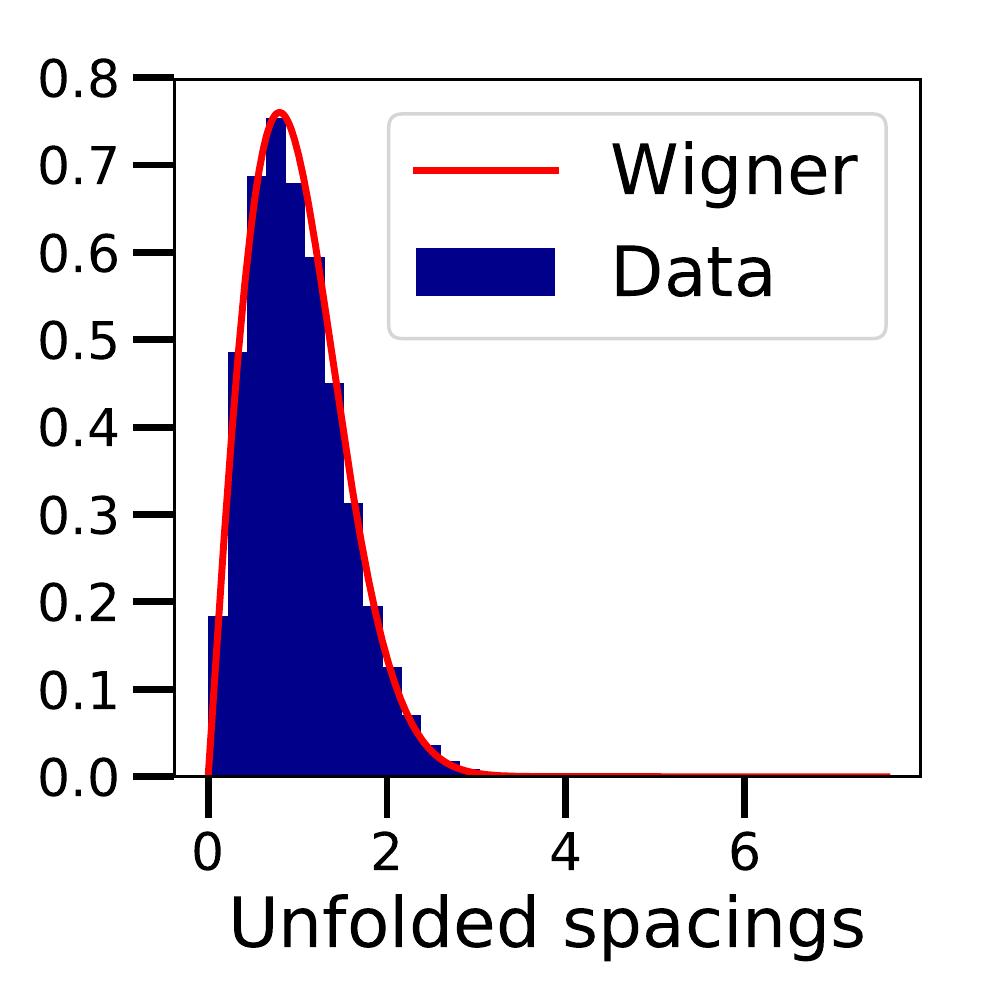}}
& \subfloat[Spacing ratios. Entire dataset.]{\includegraphics[width=0.4\textwidth]{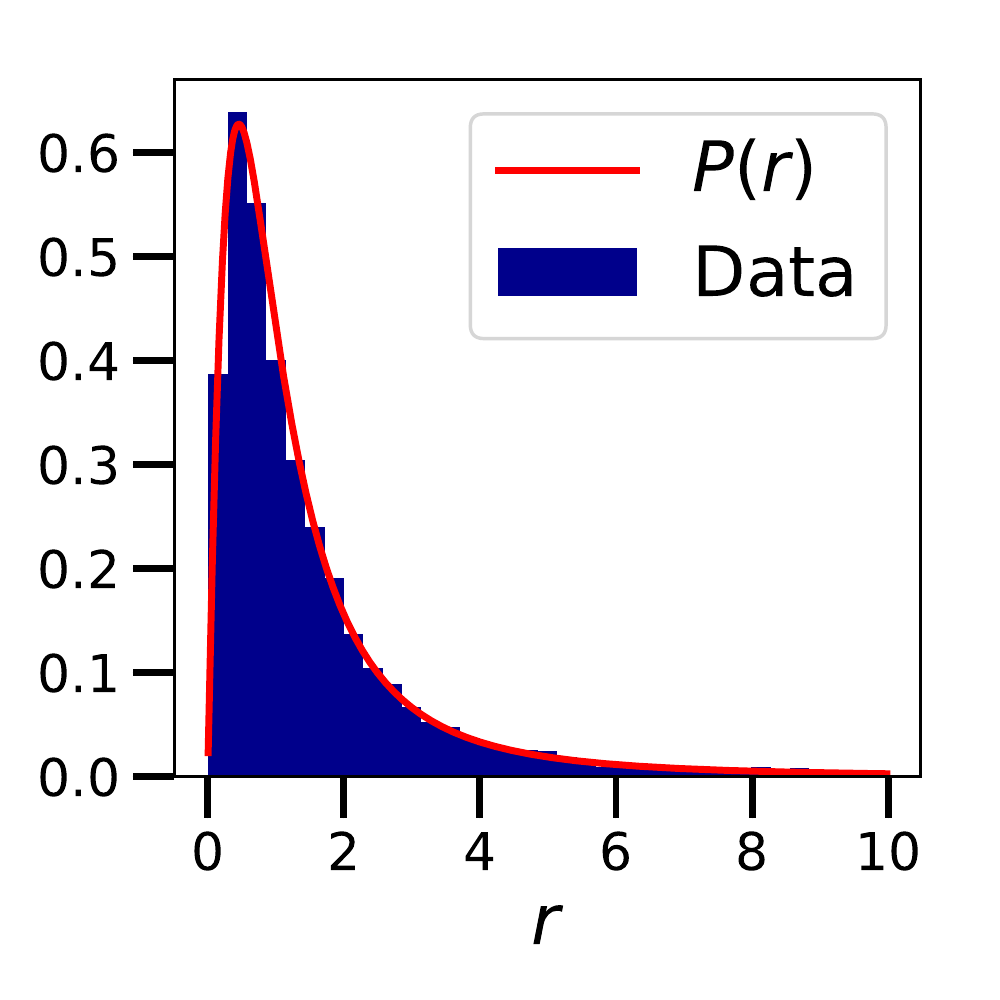}}
\end{tabular}
\caption{Spacing distributions for the Hessian of a logistic regression trained Resnet-$34$ embeddings of CIFAR10. Hessians computed over the test set.}
\label{fig:cifar_resnet_spacings}
\vspace{-10pt}
\end{figure}
\begin{figure}[h!]
\centering
\begin{tabular}{cc}
\subfloat[Unfolded spacings. Batch-size 64.]{\includegraphics[width=0.4\textwidth]{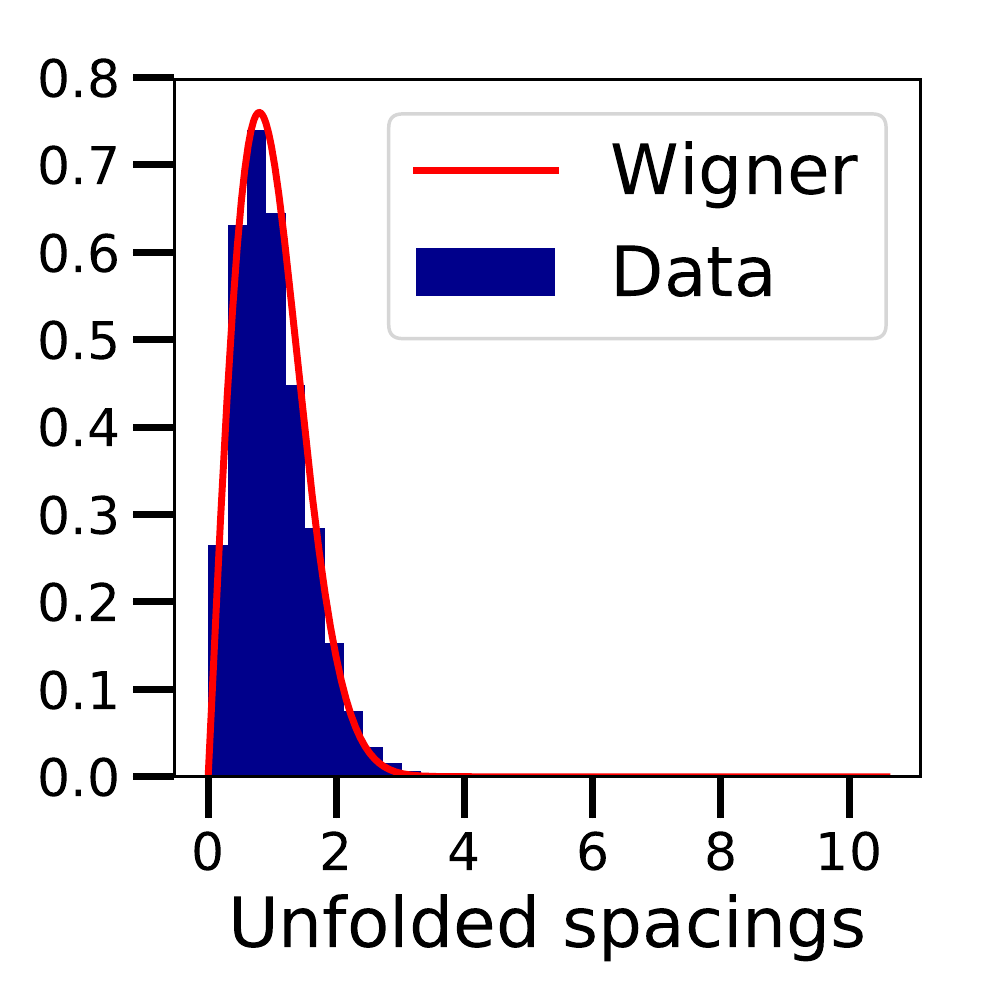}}
& \subfloat[Spacing ratios. Batch-size 64.]{\includegraphics[width=0.4\textwidth]{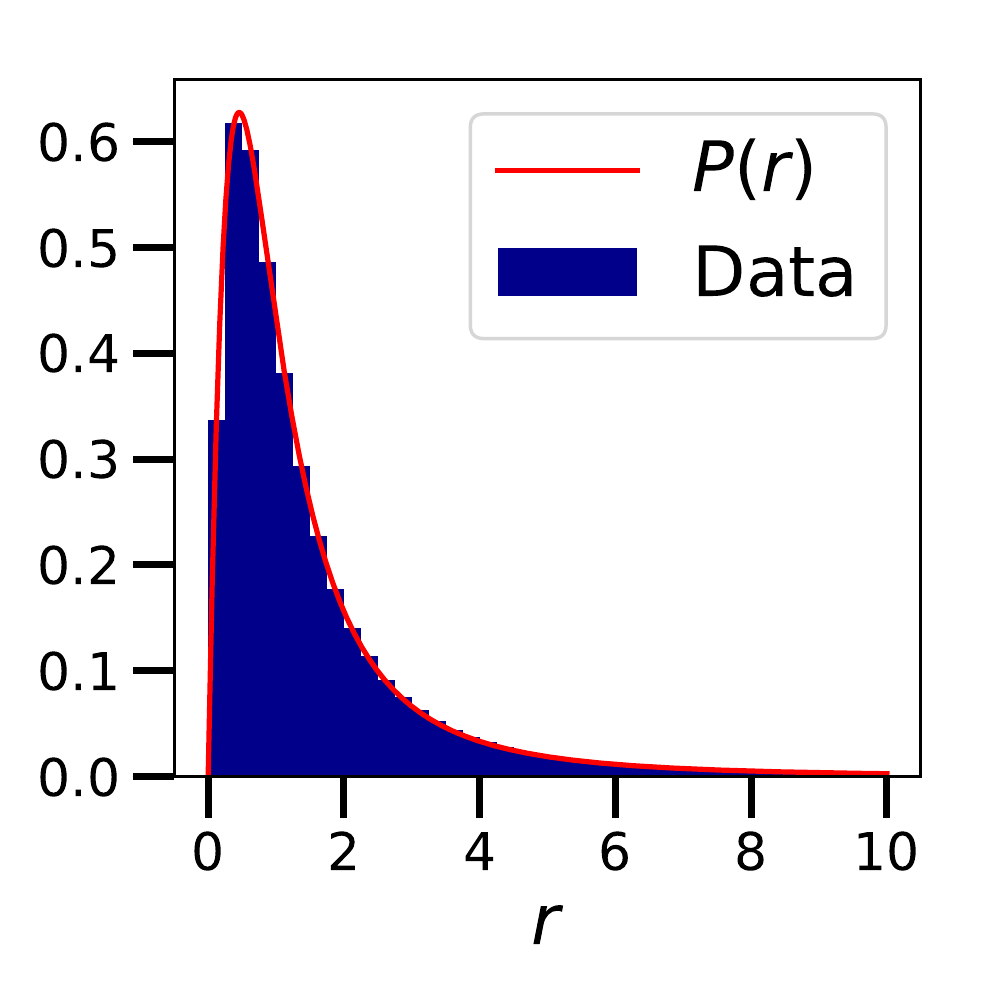}}
\end{tabular}
\caption{Spacing distributions for the Hessian of a 3-hidden-layer MLP trained on MNIST. Hessians computed over the test set.}
\vspace{-10pt}
\label{fig:deep_mlp_spacings}
\end{figure}

\physica{\subsection{MNIST and MLPs}}
We show results in Figures \ref{fig:cifar_resnet_spacings} and \ref{fig:deep_mlp_spacings}, with further plots in the supplementary material. We also considered randomly initialised networks and we evaluated the Hessians over train and test datasets separately in all cases. Unfolded spacings were computed only for Hessians evaluated on batches of 64 data points, while spacing ratios were computed in batches and over the entire dataset.
We observe a striking level of agreement between the observed spectra and the GOE. There was no discernible difference between the train and test conditions, nor between batch and full dataset conditions, nor between trained and untrained models. Note that the presence of GOE statistics for the untrained models is not a foregone conclusion. Of course, the weights of the model are indeed random Gaussian, but the Hessian is still a function of the data set, so it is not the case the Hessian eigenvalue statistics are bound to be GOE a priori. Overall, the very close agreement between Random Matrix Theory predictions and our observations for several different architectures, model sizes and datasets demonstrates a clear presence of RMT statistics in neural networks.

Our results indicate that models for the loss surfaces of large neural networks should include assumptions of GOE local statistics of the Hessian,  but ideally avoid such assumptions on the global statistics. To further illustrate this point, consider a Gaussian process $\mathcal{L}_{emp}\sim \mathcal{GP}(0, k)$ where $k$ is some kernel function. Following from our Gaussian process definition, the covariance of derivatives of the empirical loss can be computed using a well-known result (see \citet{adler2009random} equation 5.5.4), e.g.
 \begin{align} 
 \nonumber
    Cov(\partial_i \mathcal{L}_{emp}(\vec{w}), \partial_j\mathcal{L}_{emp}(\vec{w}') ) = \partial_{w_i}\partial_{w'_j} k(\vec{w}, \vec{w}')
\end{align}
and further, assuming a stationary kernel $k(\vec{w}, \vec{w}') = k\left(-\frac{1}{2}||\vec{w} - \vec{w}'||_2^2\right)$ (note abuse of notation) \begin{align}\label{eq:grad_gp_covar}
\nonumber
    & Cov(\partial_i \mathcal{L}_{emp}(\vec{w}), \partial_j\mathcal{L}_{emp}(\vec{w}') )  \\ 
    & = (w_i - w'_i)(w'_j - w_j) k''\left(-\frac{1}{2}||\vec{w} - \vec{w}'||_2^2\right)  + \delta_{ij}k'\left(-\frac{1}{2}||\vec{w} - \vec{w}'||_2^2\right). 
\end{align}

\begin{figure*}[!t]
\centering
\begin{tabular}{ccccc}
\subfloat[$k''(0)=10^{-4}$]{\includegraphics[width=0.3\textwidth]{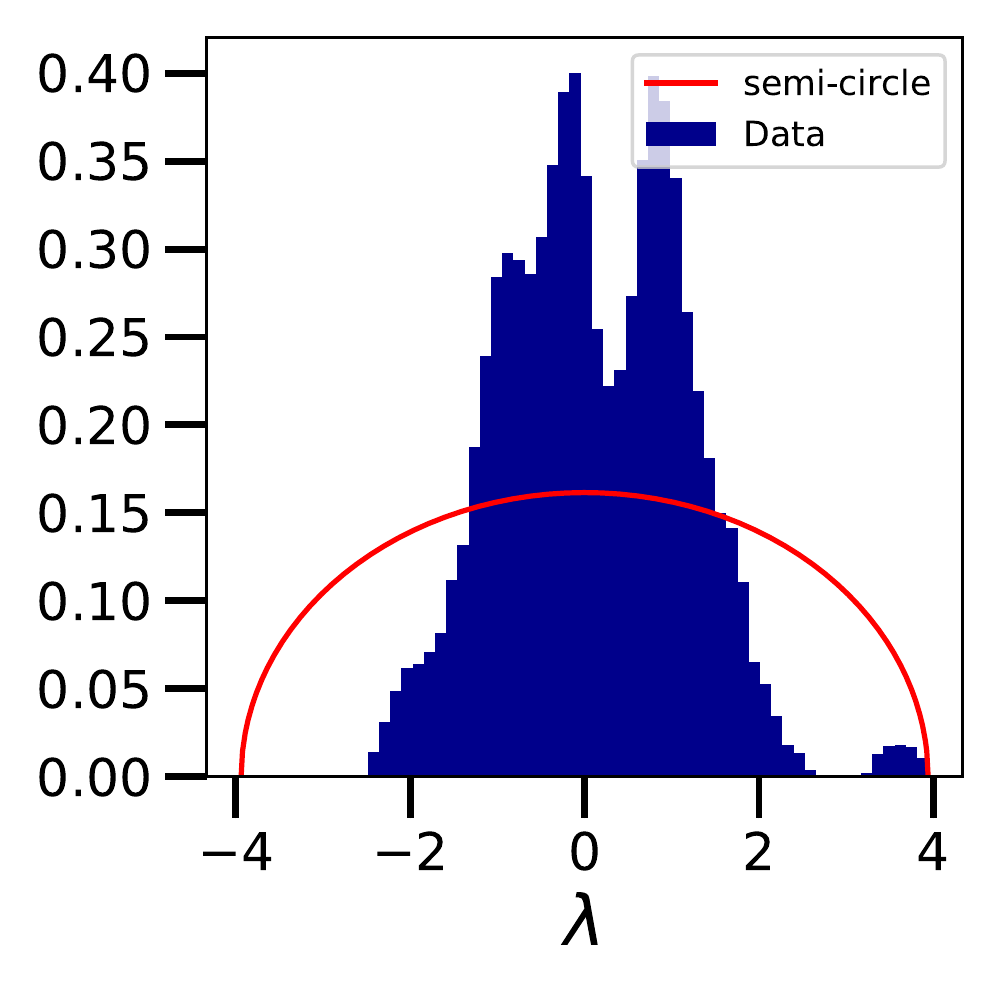}} & 
\subfloat[$k''(0)=10^{-3}$]{\includegraphics[width=0.3\textwidth]{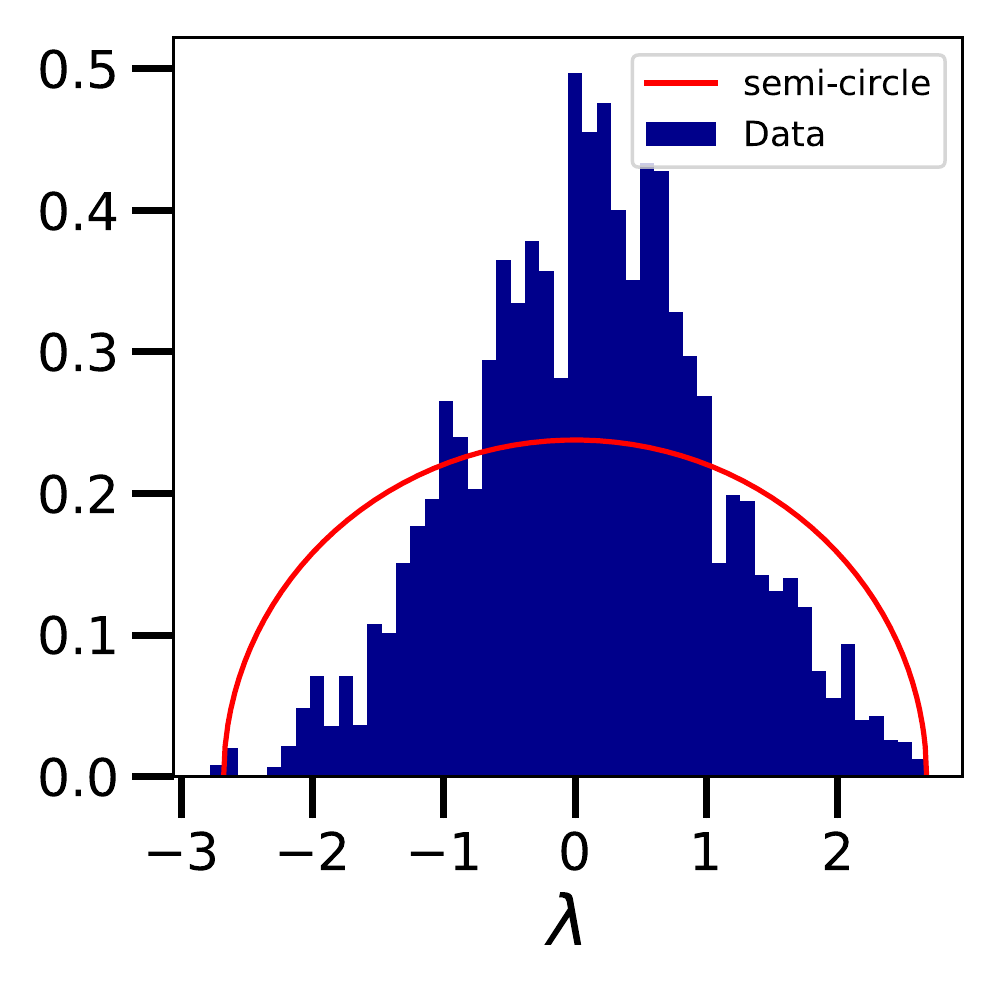}} & 
\subfloat[$k''(0)=10^{-1}$]{\includegraphics[width=0.3\textwidth]{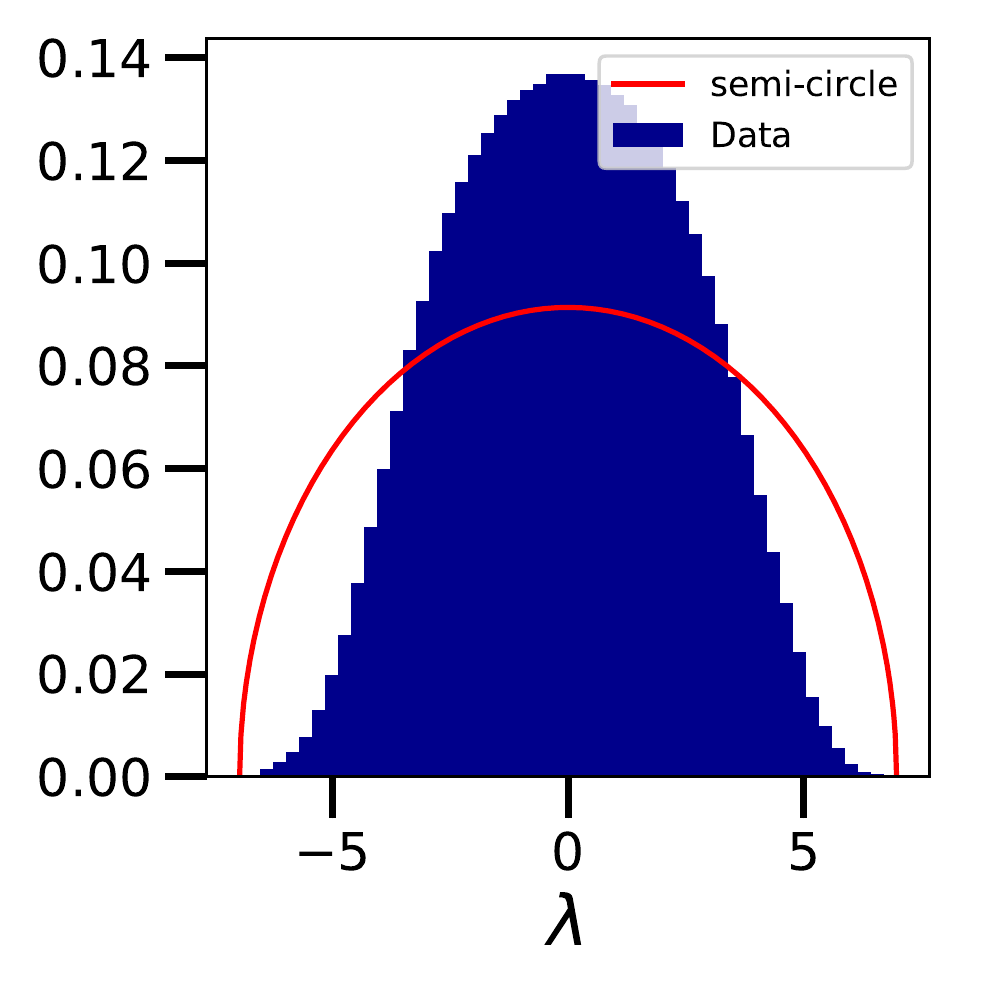}} \\
\subfloat[$k''(0)=10$]{\includegraphics[width=0.3\textwidth]{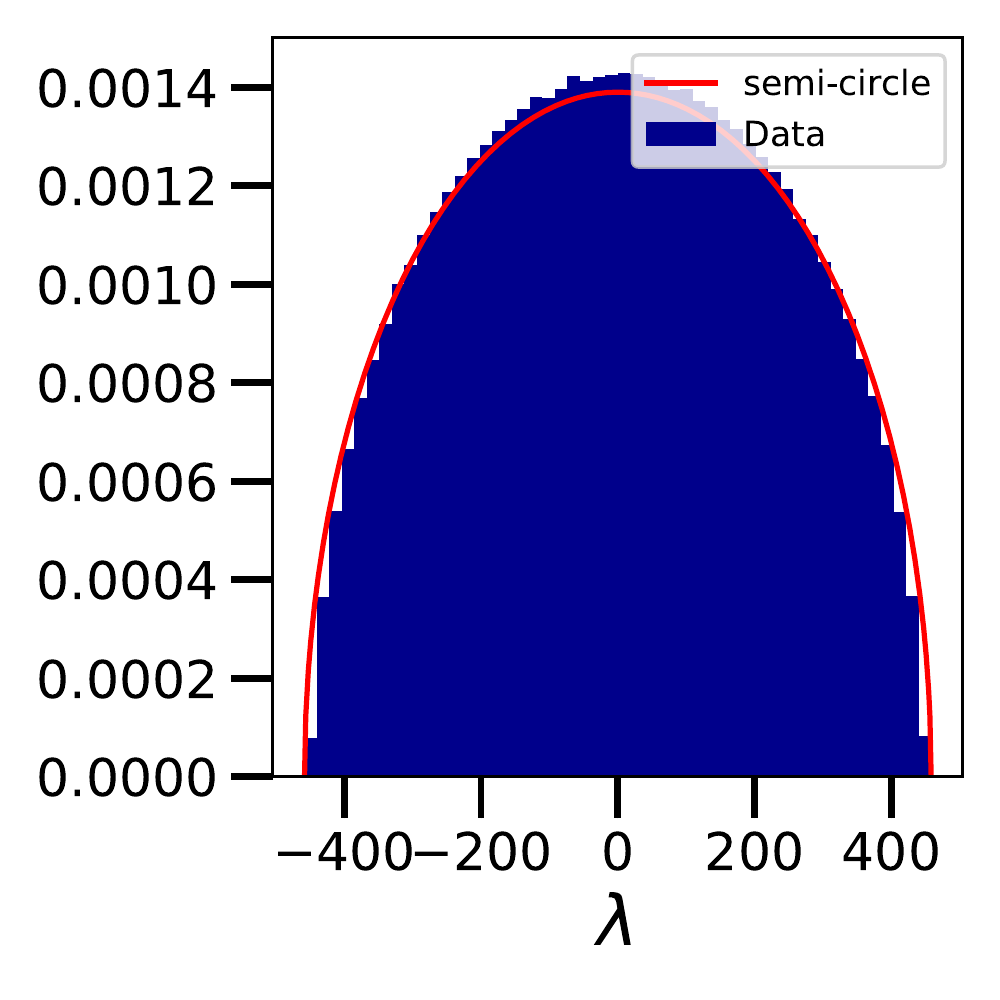}} & 
\subfloat[$k''(0)=10^{-3}*$]{\includegraphics[width=0.3\textwidth]{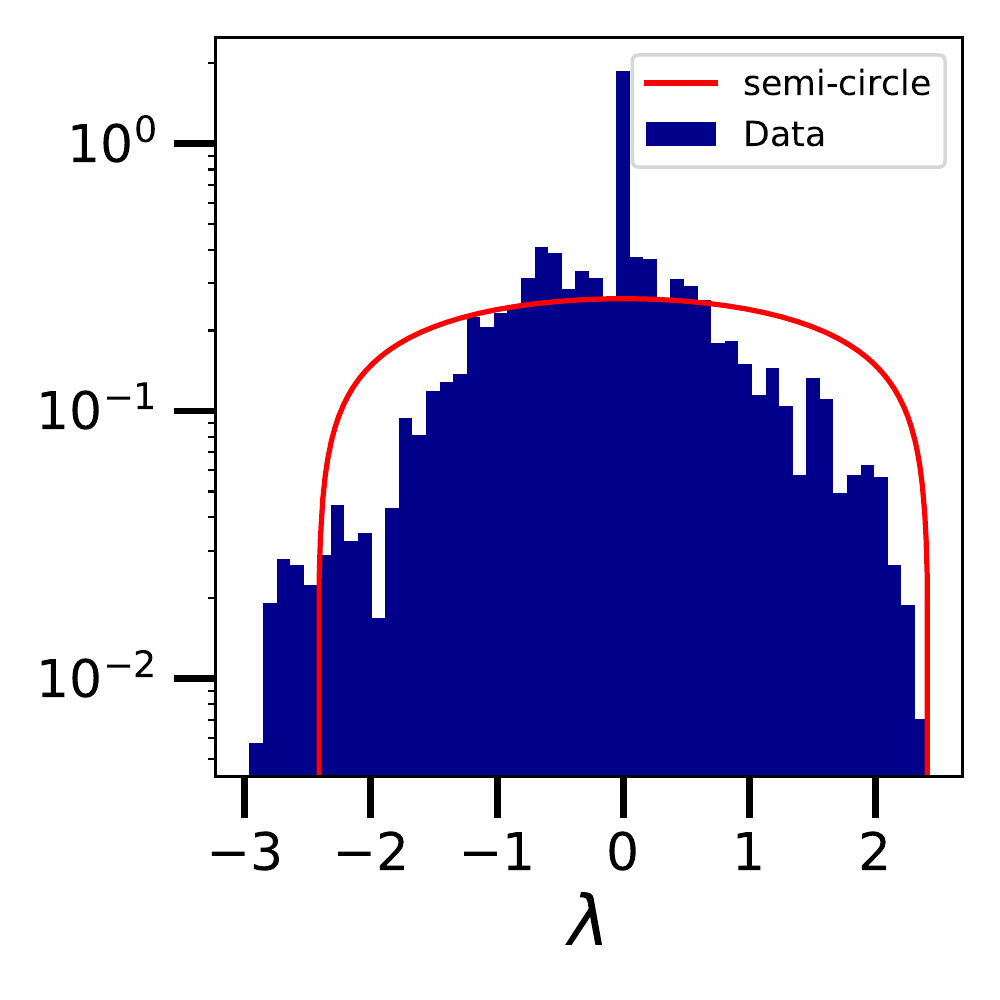}} & 
\subfloat[$k''(0)=0.0001\dagger$]{\includegraphics[width=0.3\textwidth]{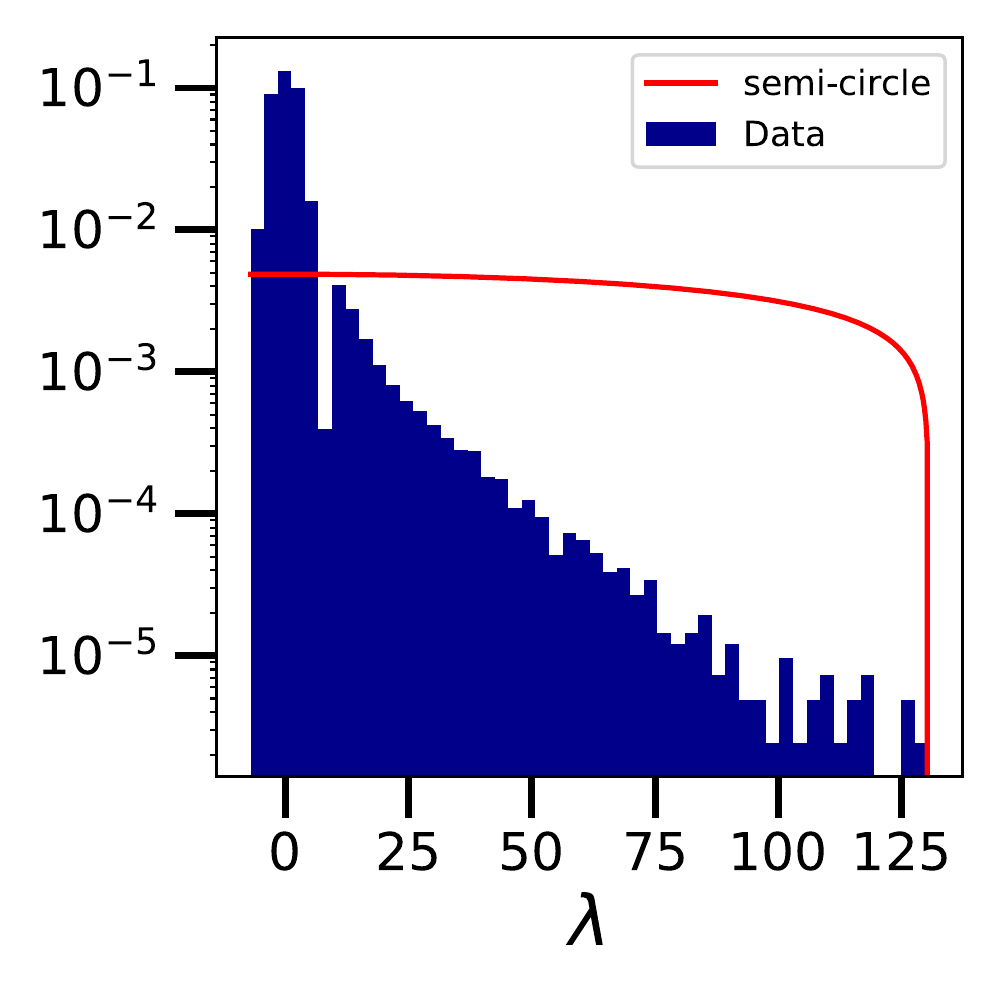}} 
\vspace{-5pt}
\end{tabular}
\caption{Spectral densities of Gaussian process Hessians with various kernel choices. All use $k'(0)=1$. The dimension is $300$ in all cases except (d), in which the Hessian is padded to 400 dimensions with zeros. All histograms are produced with 100 independent Hessian samples. $* = 100$ degenerate directions. $\dagger = 20$ outliers}
\label{fig:gp_kernel_densities}
\end{figure*}

\begin{figure*}[!t]
\centering
\vspace{-5pt}
\begin{tabular}{ccccc}
\subfloat[$k''(0)=10^{-4}$]{\includegraphics[width=0.3\textwidth]{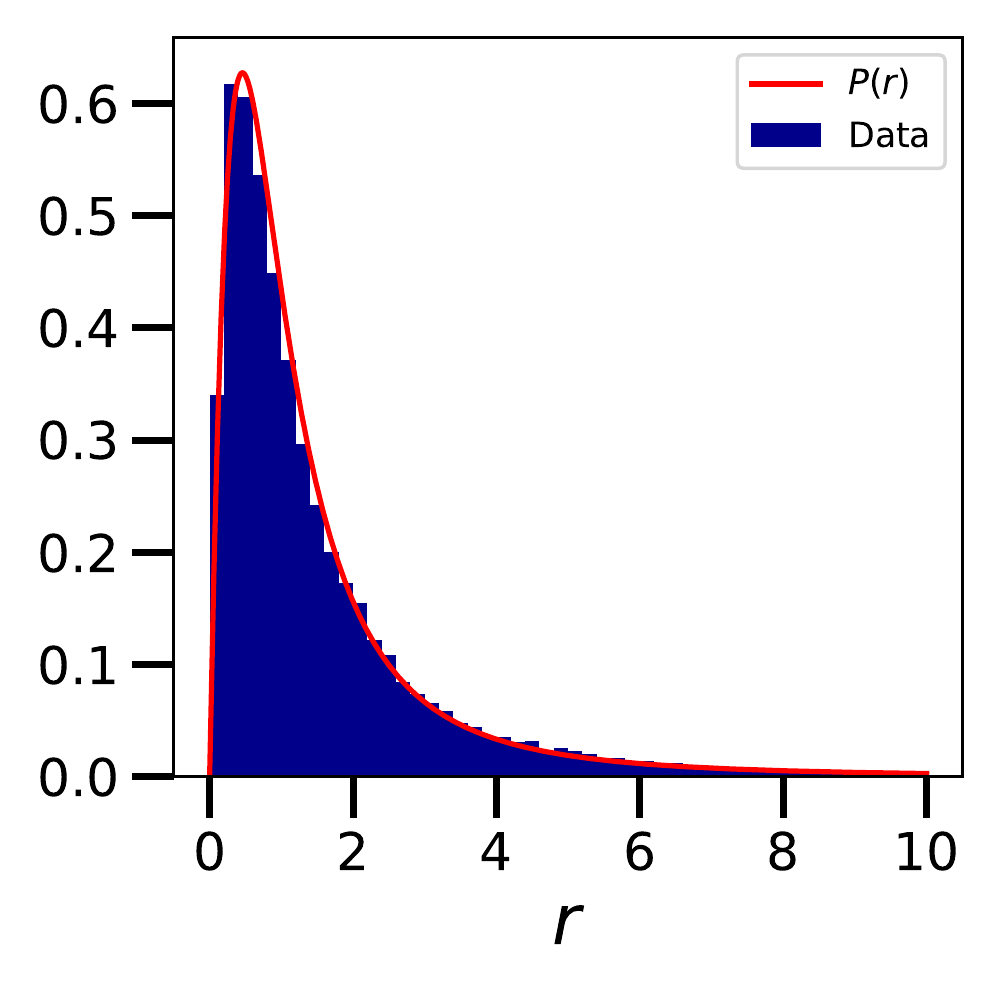}} &
\subfloat[$k''(0)=10^{-3}$]{\includegraphics[width=0.3\textwidth]{_5526544112871594660.pdf}} & 
\subfloat[$k''(0)=10^{-1}$]{\includegraphics[width=0.3\textwidth]{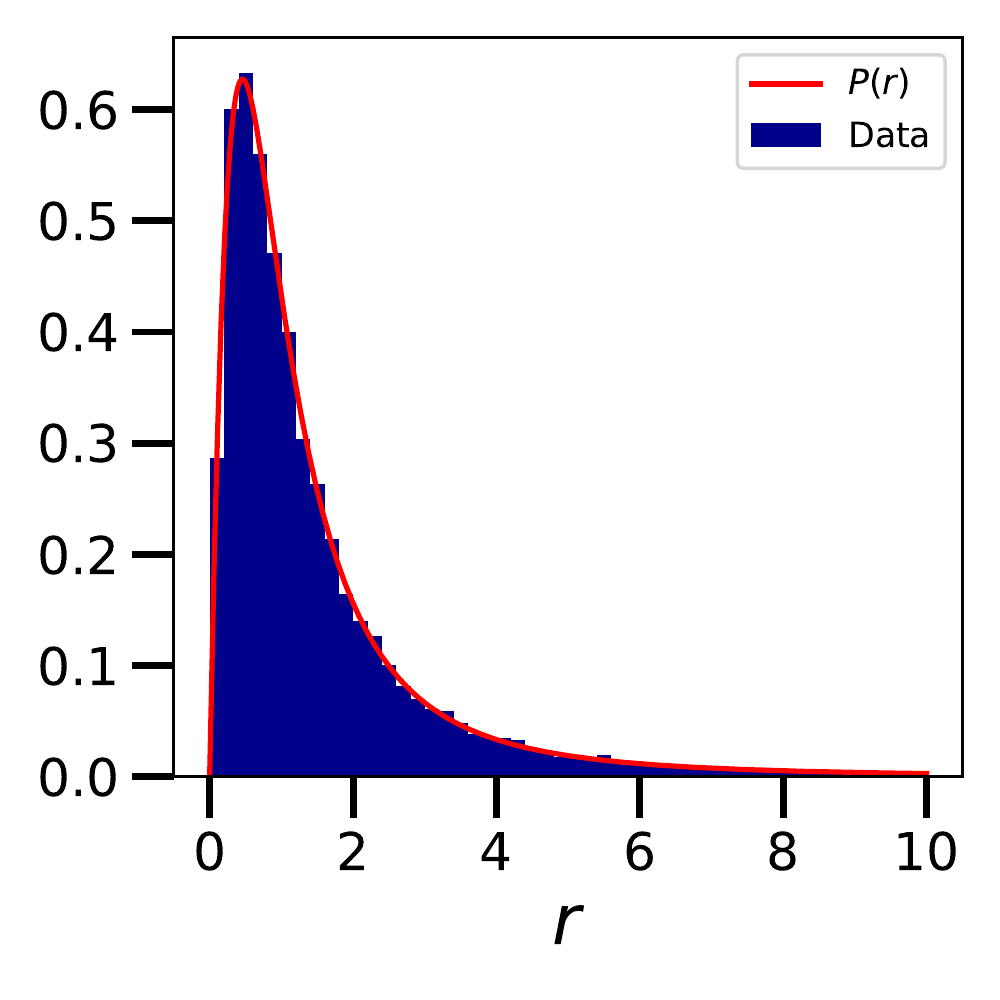}} \\
\subfloat[$k''(0)=10$]{\includegraphics[width=0.3\textwidth]{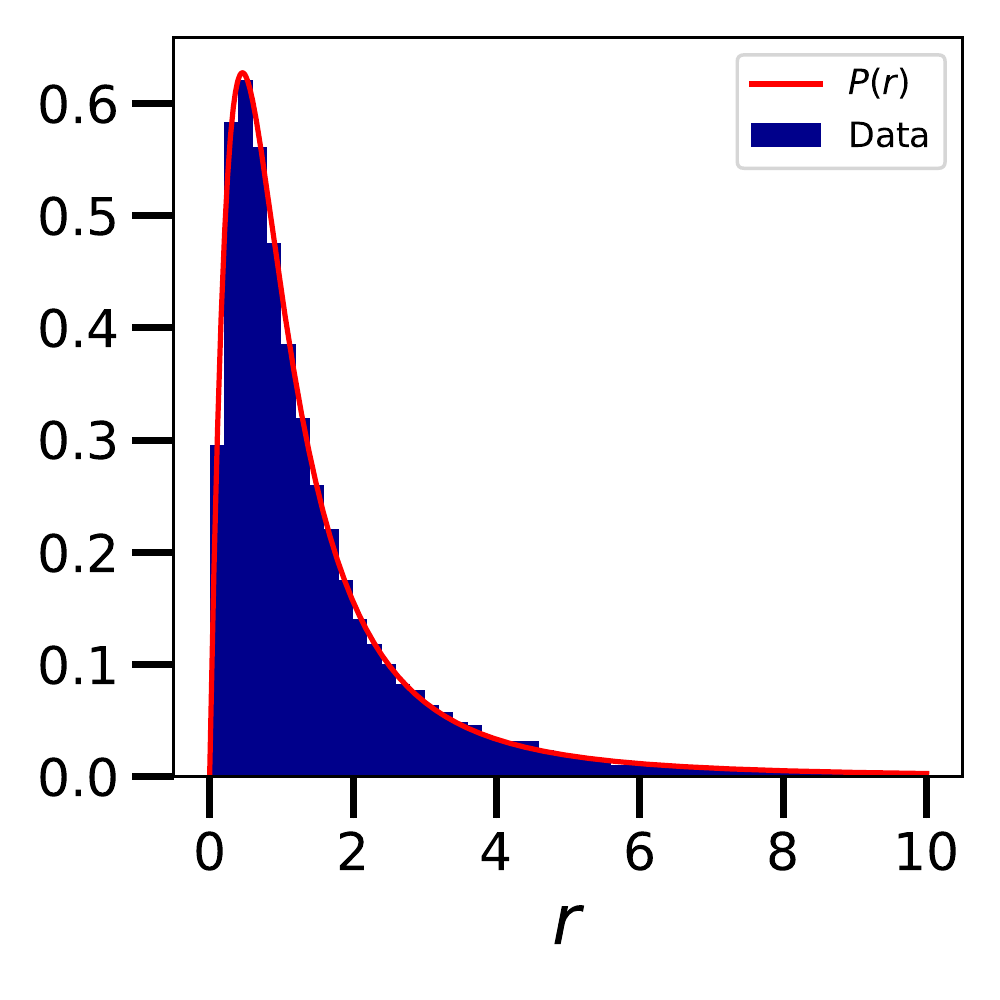}} & 
\subfloat[$k''(0)=10^{-3}*$]{\includegraphics[width=0.3\textwidth]{_5526544112871594660.pdf}} & 
\subfloat[$k''(0)=0.0001\dagger$]{\includegraphics[width=0.3\textwidth]{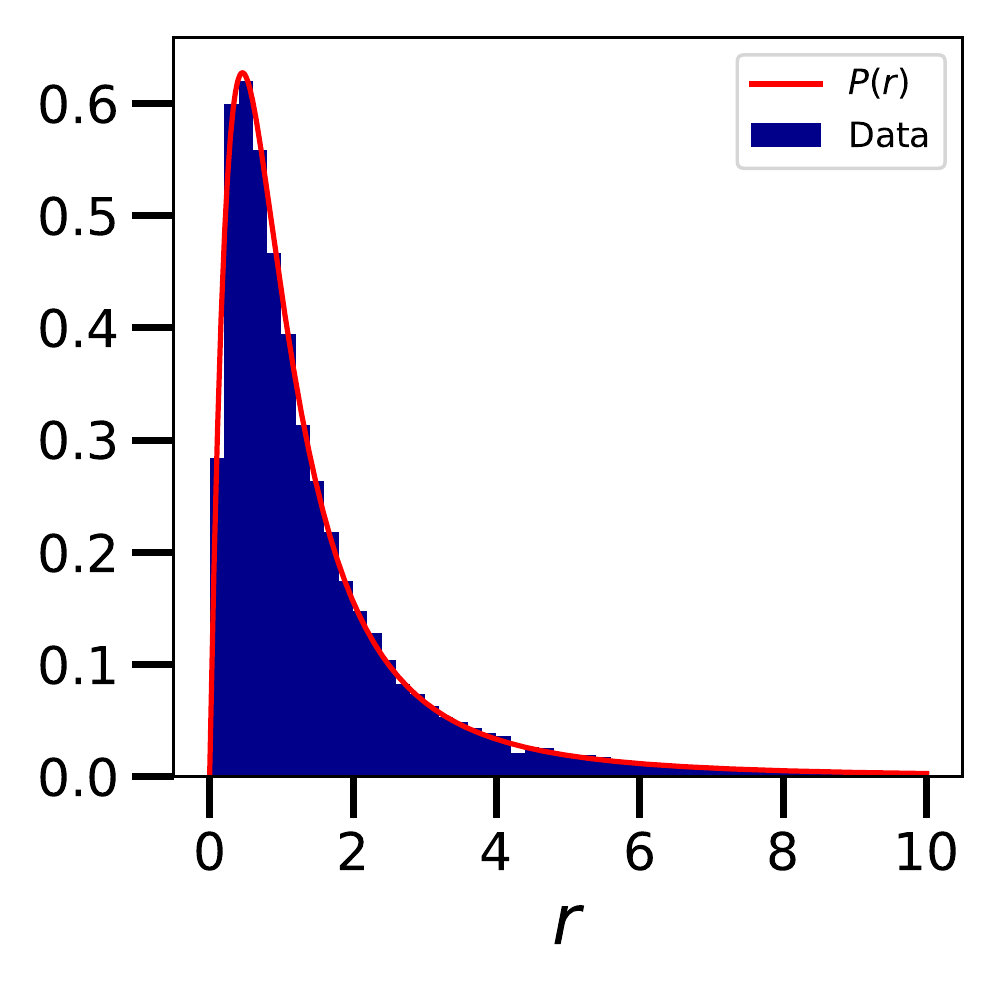}} 
\end{tabular}
\caption{Consecutive spacing ratios of Gaussian process Hessians with various kernel choices. All use $k'(0)=1$. The dimension is $300$ in all cases except (d), in which the Hessian is padded to 400 dimensions with zeros. $* = 100$ degenerate directions. $\dagger = 20$ outliers.}
\vspace{-5pt}
\label{fig:gp_kernel_ratios}
\end{figure*}

Differentiating (\ref{eq:grad_gp_covar}) further, we obtain 
\begin{align}
   & Cov(\partial_{ij}\mathcal{L}_{emp}(\vec{w}), \partial_{kl} \mathcal{L}_{emp}(\vec{w}))  = k''(0)\left(\delta_{ik}\delta_{jl} + \delta_{il}\delta_{jk}\right) + k'(0)^2 \delta_{ij}\delta_{kl}
\end{align}
The Hessian $\mH_{emp}$ has Gaussian entries with mean zero, so the distribution of $\mH_{emp}$ is determined entirely by $k'(0)$ and $k''(0)$. Neglecting to choose $k$ explicitly, we vary the values of $k'(0)$ and $k''(0)$ to produce nearest neighbour spectral spacings ratios and spectral densities. The histograms for spectral spacing ratios are indistinguishable and agree very well with the GOE, as shown in Figure \ref{fig:gp_kernel_ratios}. The spectral densities are shown in Figure \ref{fig:gp_kernel_densities}, including examples with rank degeneracy, introduced by defining $k$ only on a lower-dimensional subspace of the input space, and outliers, introduced by adding a fixed diagonal matrix to the Hessian. Figure \ref{fig:gp_kernel_densities} shows varying levels of agreement with the semi-circle law, depending on the choice of $k'(0), k''(0)$.

\begin{figure}[h!]
\centering
\begin{tabular}{cc}
\subfloat[Mean spectral density]{\includegraphics[width=0.4\textwidth]{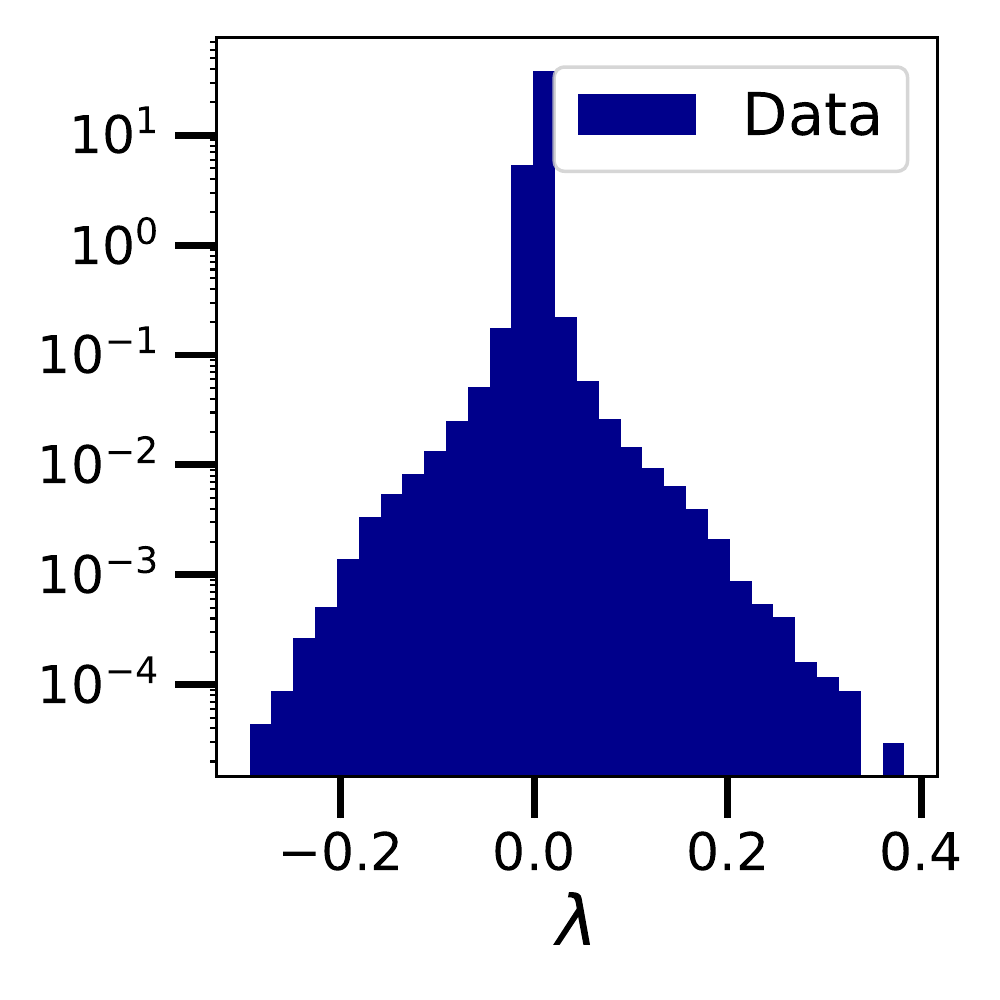}}
& \subfloat[Spacing ratios.]{\includegraphics[width=0.4\textwidth]{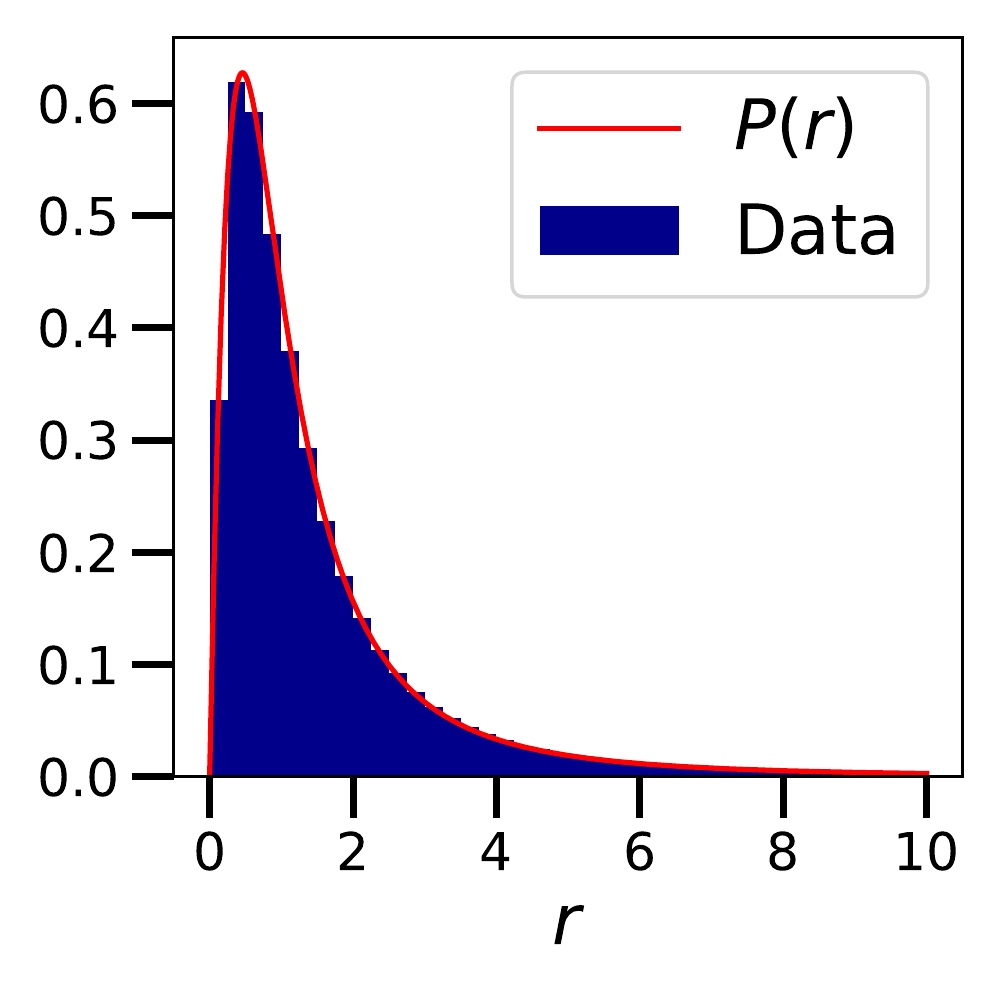}}
\end{tabular}
\caption{Spectral statistics for the Hessian of a CNN trained on CIFAR10. Hessians computed over batches of size 64 on the test set.}
\label{fig:cifar_lenet}
\vspace{-10pt}
\end{figure}

\physica{\subsection{Beyond the MLP}}
\physica{Figure \ref{fig:cifar_lenet} shows the mean spectral density and adjacent spacing ratios for the Hessian of a CNN trained on CIFAR10. As with the MLP networks and MNIST data considered above, we see an obviously non-semicircular mean level density but the adjacent spacing ratios are nevertheless described by the universal GOE law.}

\begin{figure}[h!]
\centering
\begin{tabular}{cc}
\subfloat[Mean spectral density]{\includegraphics[width=0.4\textwidth]{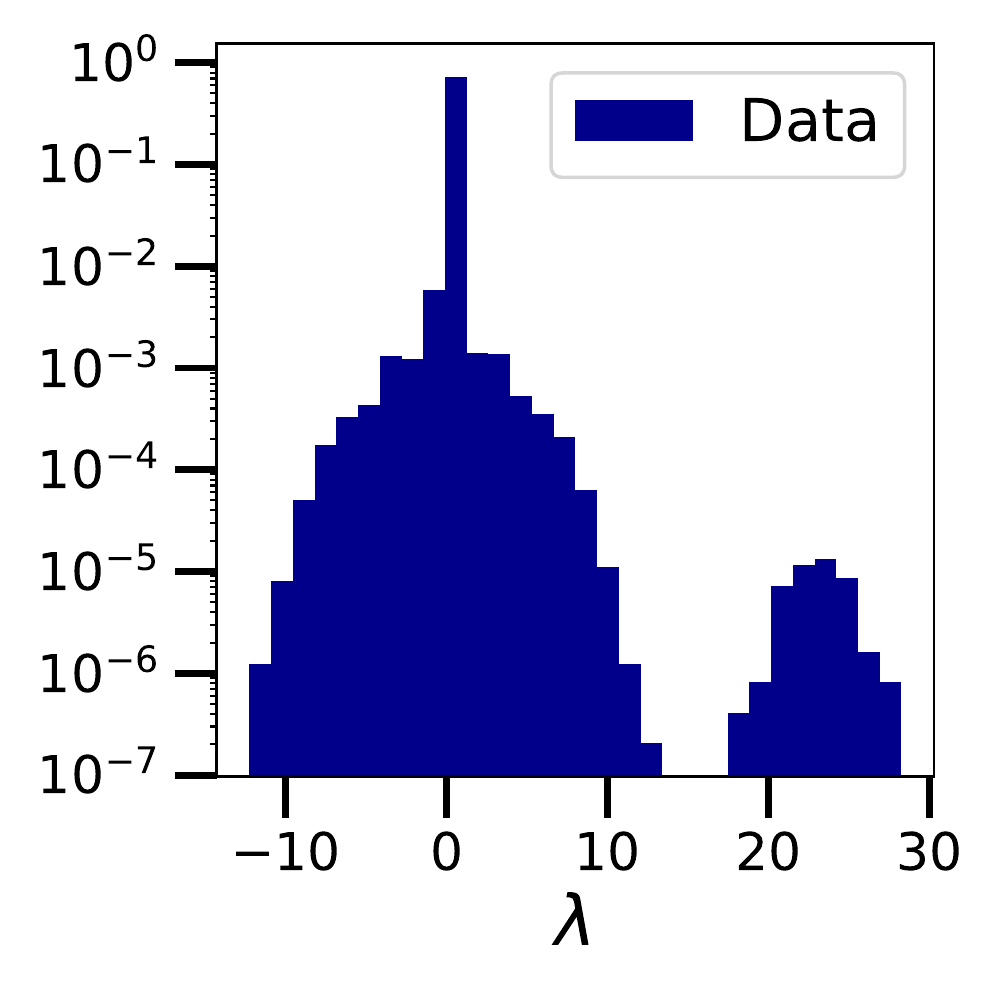}}
& \subfloat[Spacing ratios.]{\includegraphics[width=0.4\textwidth]{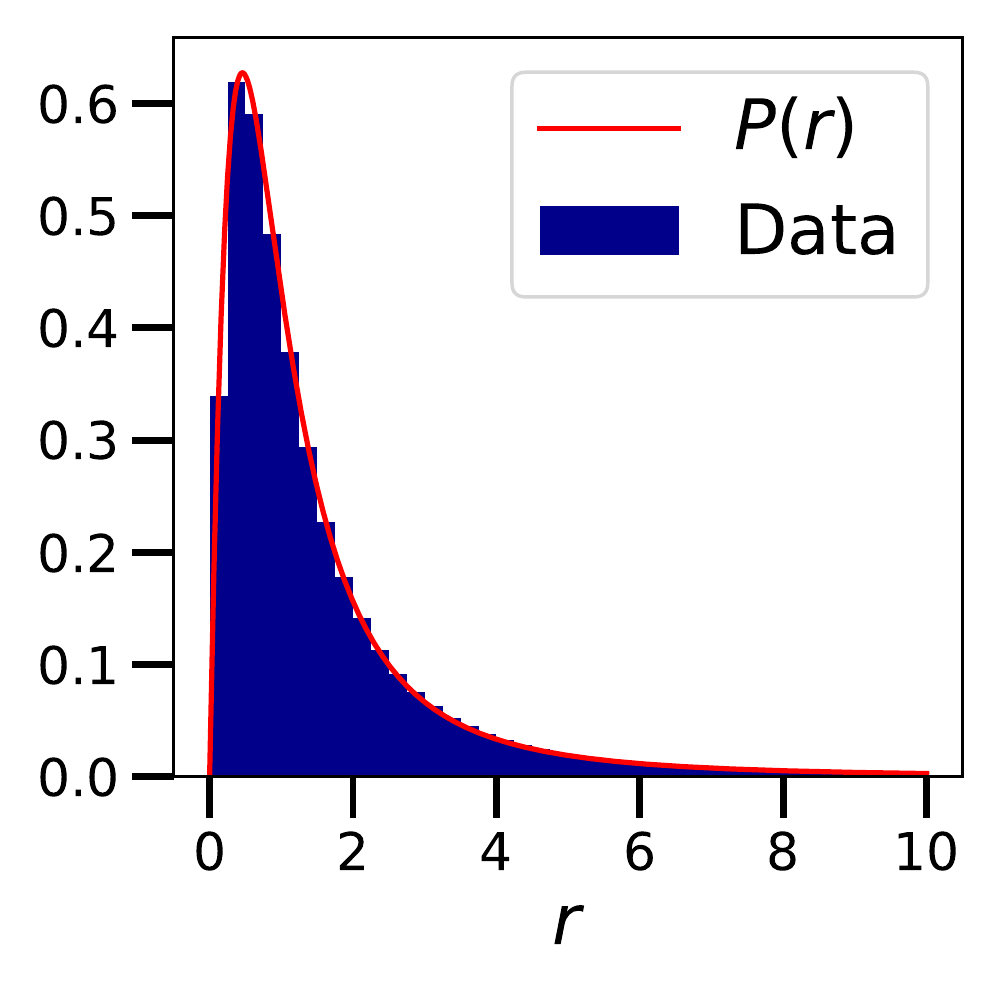}}
\end{tabular}
\caption{Spectral statistics for the Hessian of an MLP trained on the Bike dataset. Hessians computed over batches of size 64 on the test set.}
\label{fig:mlp_bike_hessian}
\vspace{-10pt}
\end{figure}

\physica{\subsection{Beyond image classification}}
\physica{Figure \ref{fig:mlp_bike_hessian} shows the mean spectral density and adjacent spacing ratios for the Hessian of an MLP trained on the Bike dataset. Once again we see an obviously non-semicircular mean level density but the adjacent spacing ratios are nevertheless described by the universal GOE law. This serves to demonstrate that there is nothing special about image data or, more importantly, high input feature dimension, since the Bike dataset has only 13 input features.}

\begin{figure}[h!]
\centering
\begin{tabular}{cc}
\subfloat[Mean spectral density]{\includegraphics[width=0.4\textwidth]{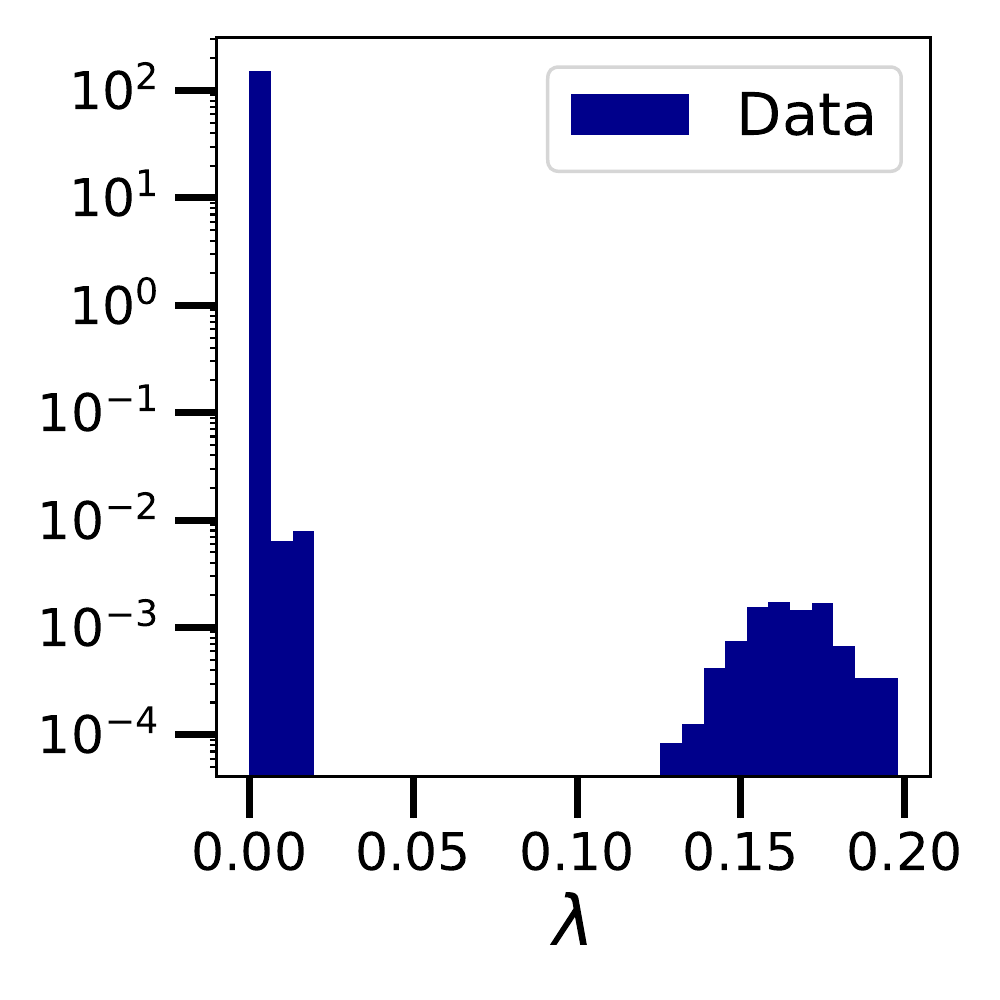}}
& \subfloat[Spacing ratios.]{\includegraphics[width=0.4\textwidth]{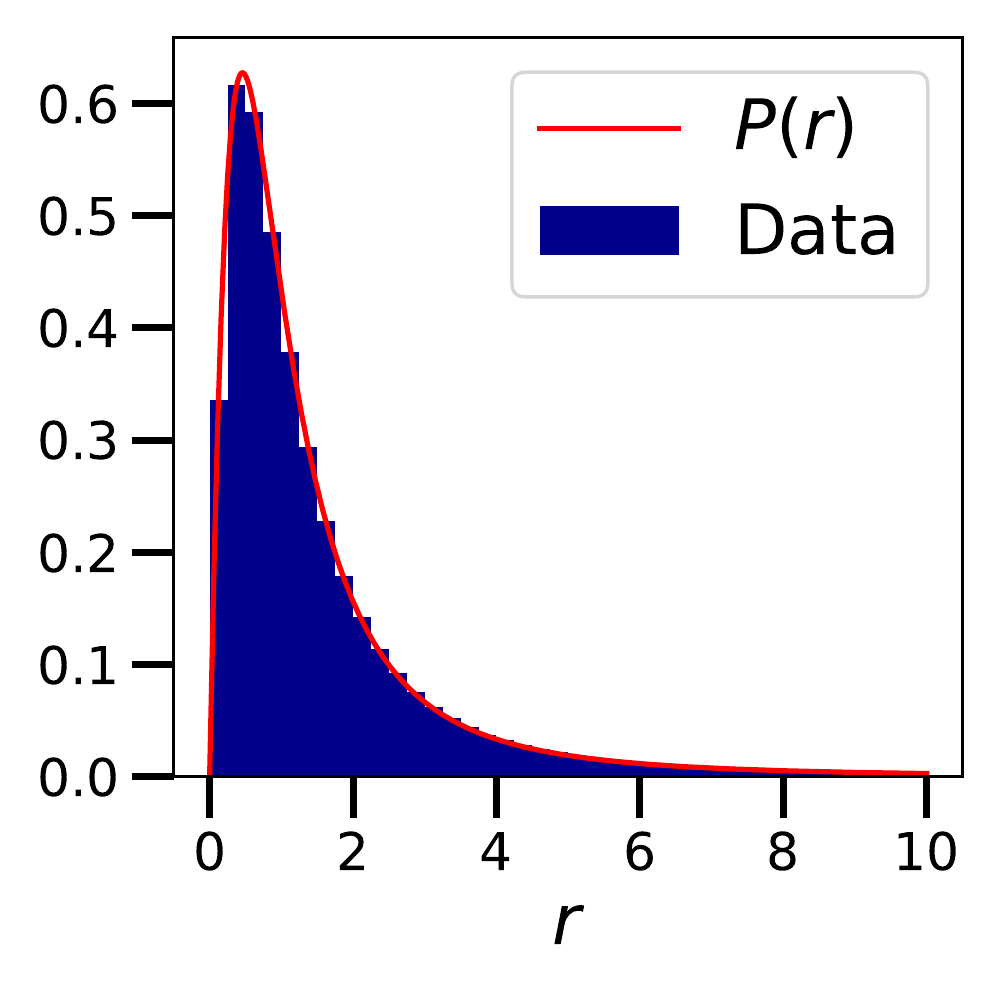}}
\end{tabular}
\caption{Spectral statistics for the Gauss-Newton matrix of an MLP trained on the Bike dataset. Matrices computed over batches of size 64 on the test set.}
\label{fig:mlp_bike_gn}
\vspace{-10pt}
\end{figure}

\physica{\subsection{Beyond the Hessian}
Given that the Hessian is not the only matrix of interest in Machine Learning, it is pertinent to study whether our empirical results hold more generally. There have been lots of investigations for the Gauss-Newton \citep{loke2002comparison,pennington2017geometry}, or generalised Gauss-Newton (which is the analogue of the Gauss-Newton when using the cross entropy instead of square loss) matrices, particularly in the fields of optimisation \citep{dauphin2014identifying,martens2012training,martens2015optimizing,martens2014new}. We consider the Gauss-Newton of the network trained on the Bike dataset with square loss. In this case the Gauss Newton $\bold{G} = \bold{J^{T}}\bold{J}$ shares the same non-null subspace as the Neural Tangent Kernel (NTK) \citep{jacot2018neural,cai2019gram}, where $\bold{J}$ denotes the Jacobian, i.e the derivative of the output with respect to the weights, which in this case is simply a vector. The NTK is used for the analysis of trajectories of gradient descent and is particularly interesting for large width networks, where it can be analytically shown that weights remain close to their initialisation and the network is well approximated by its linearisation. }
\physica{Figure \ref{fig:mlp_bike_gn} shows the mean spectral density and adjacent spacing ratios for the Gauss-Newton matrix of an MLP trained on the Bike dataset. The results are just as for the Hessians above: universal GOE spacings, but the mean density is very much not semicircular. This is an interesting result because even for a different matrix employed in a different context we still see the same universal RMT spacings.}

\section{Conclusion and future work}
We have demonstrated experimentally the existence of random matrix statistics in small neural networks on the scale of the mean eigenvalue separation. This provides the first  direct evidence of universal RMT statistics present in neural networks trained on real datasets. \physica{Hitherto the role of random matrix theory in deep learning has been unclear. Prior work has studied theoretical models with specific assumptions leading to specific random matrix ensembles. Though certainly insightful, it is not clear to what extent any of these studies are applicable to real neural networks. This work aims to shift the focus by demonstrating the clear presence of universal random matrix behaviour in real neural networks. We expect that future theoretical studies will start from this robust supposition.} 

When working with a neural network on some dataset, one has information a priori about its Hessian. \physica{Its distribution and correlation structure may well be entirely inaccessible, but correlations between Hessian eigenvalues on the local scale can be assumed to be universal and overall the matrix can be rightly viewed as a random matrix possessing universal local statistics.} 

We focus on small neural networks where Hessian eigendecomposition is feasible. Future research that our work motivates could develop methods to approximate the level spacing distribution of large deep neural networks for which exact Hessian spectra cannot be computed. If the same RMT statistics are found, this would constitute a profound universal property of neural networks models; conversely, a break-down in these RMT statistics would be an indication of some fundamental separation between different network sizes or architectures.

\physica{A few recent works \cite{loureiro2021capturing, goldt2020gaussian, adlam2020neural} considered and used the idea of \emph{Gaussian equivalence} to make theoretical progress in neural network models with fewer assumptions than previously required (e.g. on the data distribution). The principle is that complicated random matrix distributions on non-linear functions of random matrices can be replaced in  calculations training and test loss by their Gaussian equivalents, i.e. Gaussian matrices with matching first and second moments. This idea reflects a form of universality and can drastically increase the tractability of calculations. The random matrix universality we have here demonstrated in neural networks may be related, and should be considered as a possible source of other analogous universality simplifications that can render realistic but intractable models tractable.}

One intriguing possible avenue is the relation to chaotic systems. Quantum systems with chaotic classical limits are know to display RMT spectral pairwise correlations, whereas Poisson statistics correspond to integrable systems. We suggest that the presence of GOE pairwise correlations in neural network Hessians, as opposed to Poisson, indicates that neural network training dynamics cannot be reduced to some simpler, smaller set of dynamical equations.

\section*{Acknowledgements}
JPK is pleased to acknowledge support from ERC Advanced Grant 740900 (LogCorRM). DMG is grateful for the support from the JADE computing facility and in particular the extensive support of Andrew Gittings. NPB is grateful for the support of the Advanced Computing Research Centre of the University of Bristol. Furthermore the authors would like to thank Samuel Albanie for extensive discussions on the exponential hardness of the true loss.

\appendix
\section{Extra Figures and Degeneracy Investigation}\label{sec:degen}
Figure \ref{fig:cifarresnet_comp_mnist_truncation} compares the effect of degeneracy on unfolded spacings in each of the 3 cases considered. We see that the logistic MNIST models (trained and untrained) have a much greater level of degeneracy, whereas the CIFAR10-Resnet34 spectra clearly have GOE spacings even without any cut-off. Figures \ref{fig:log_mnist_spacings_ratio}–\ref{fig:log_mnist_unt_spacings_ratio} show further unfolded spacing and spacing ratio results like those in the main text.
\begin{figure}[h]
\centering
\begin{tabular}{cc}
\subfloat[Batch train]{\includegraphics[width=0.4\textwidth]{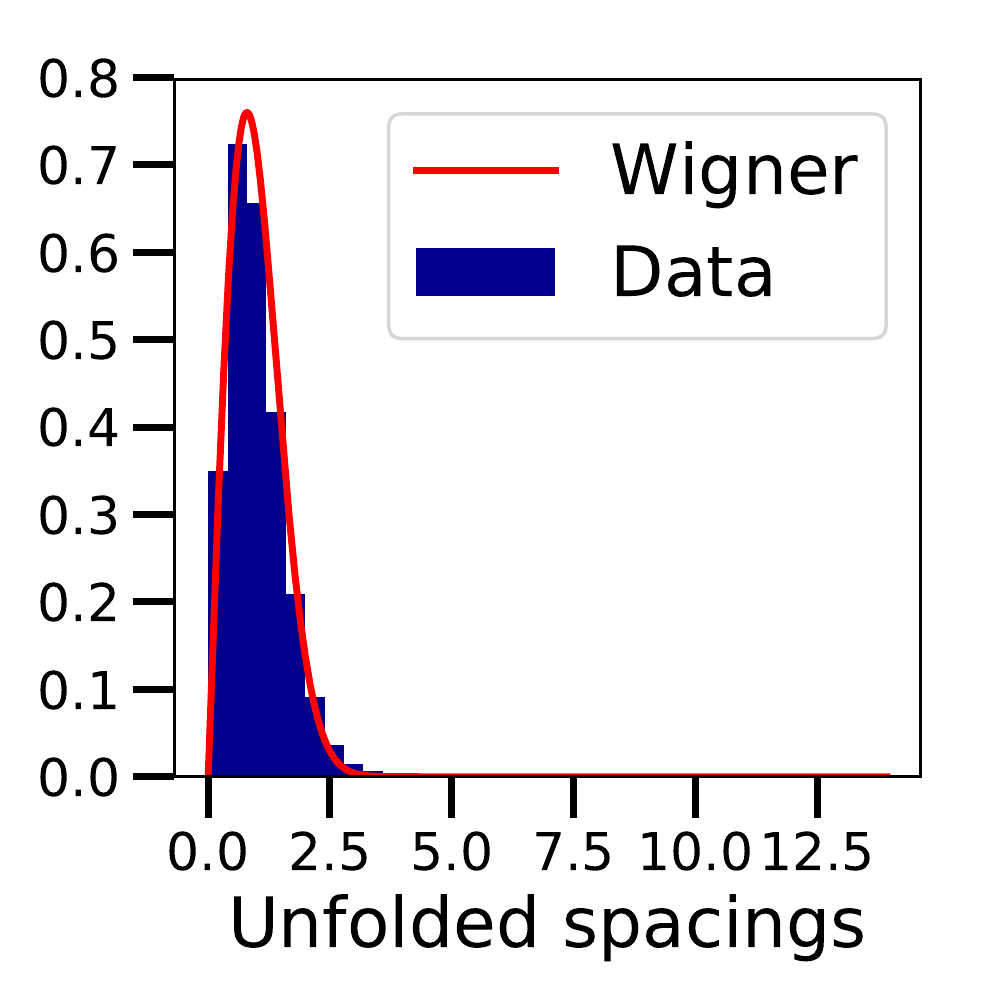}}
& \subfloat[Batch test]{\includegraphics[width=0.4\textwidth]{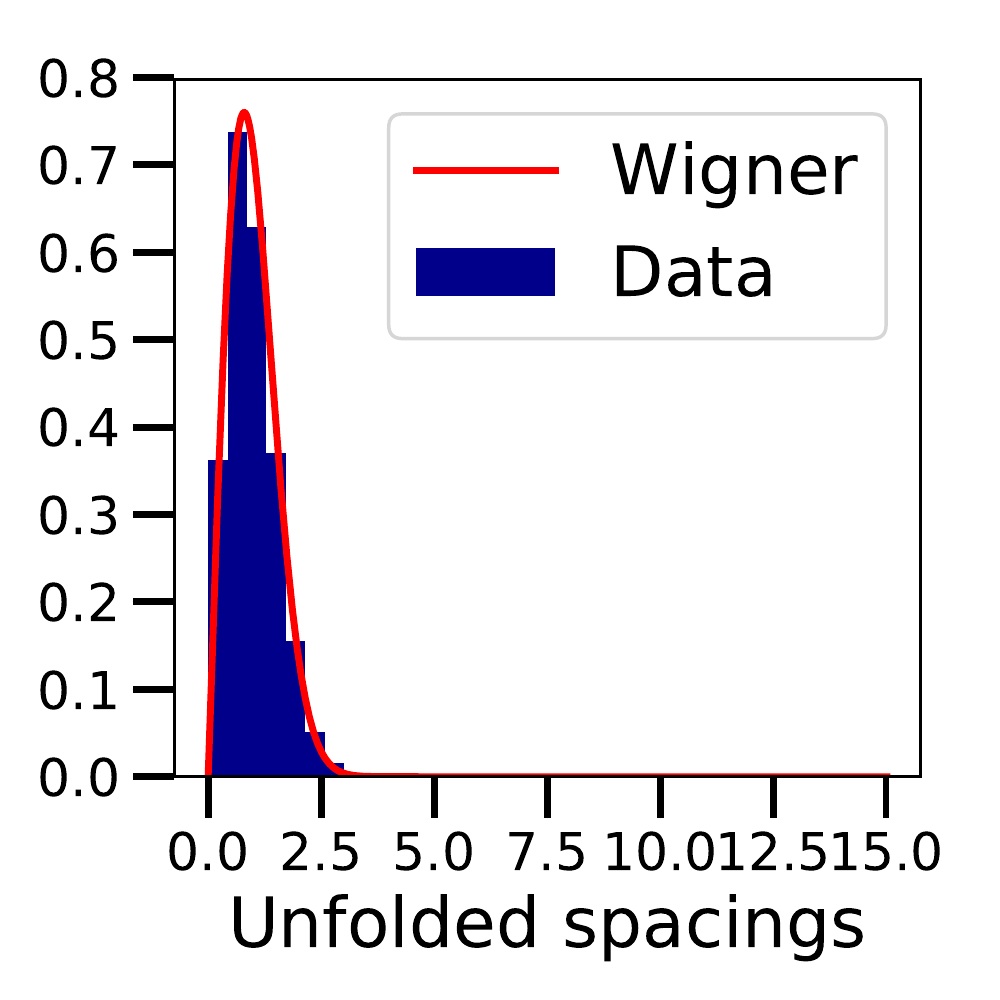}}
\end{tabular}
    \caption{Unfolded spacings for the Hessian of a logistic regression trained on MNIST. Hessian computed batches of size 64 of the training and test datasets.}
    \label{fig:log_mnist_spacings_unfolded}
\end{figure}

\begin{figure}[h]
\centering
\begin{tabular}{cccc}
\subfloat[All train]{\includegraphics[width=0.4\textwidth]{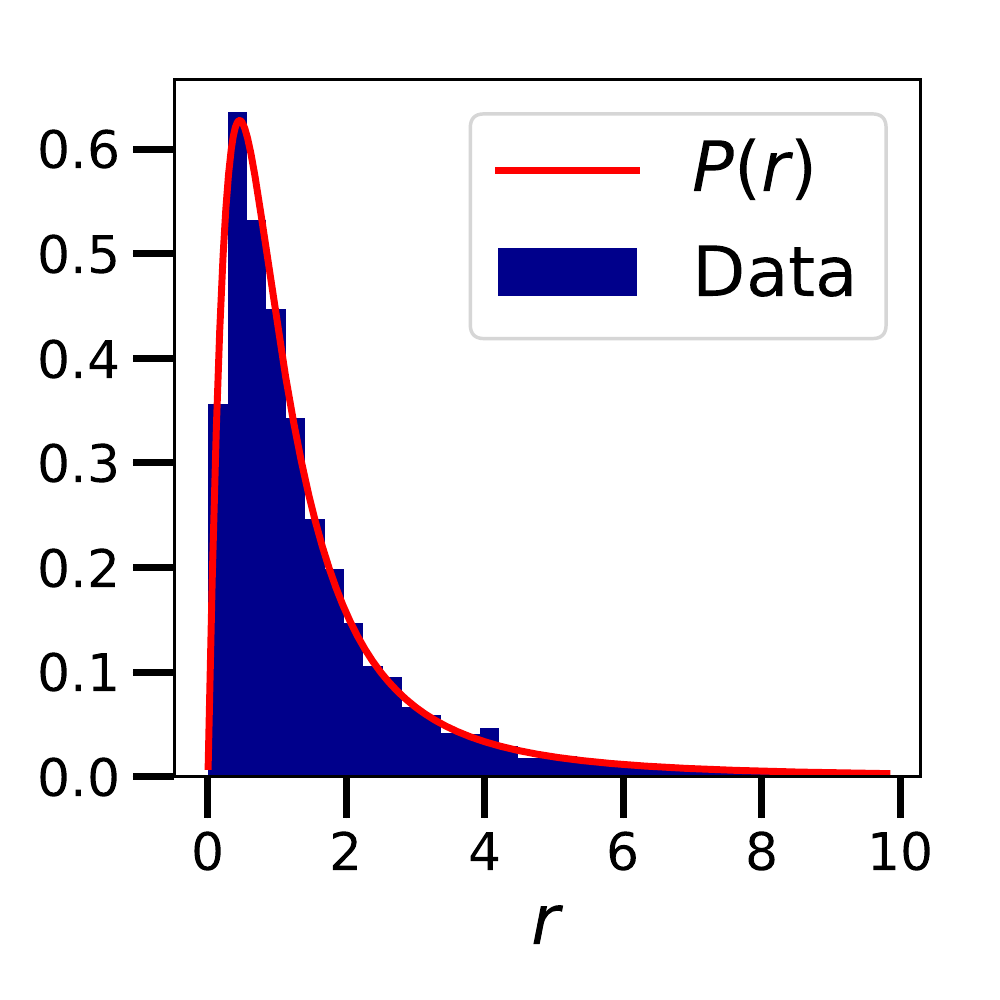}}
& \subfloat[All test]{\includegraphics[width=0.4\textwidth]{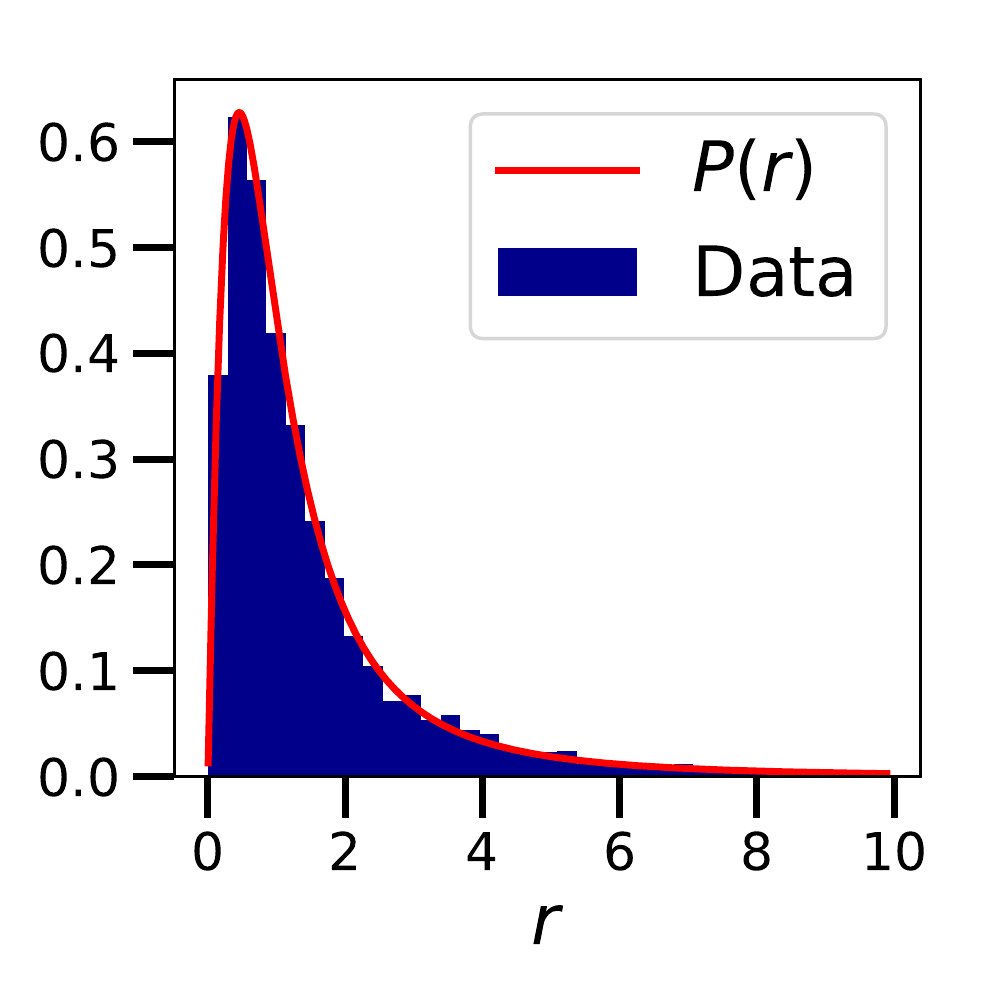}}
\end{tabular}
\caption{Consecutive spacing ratios for the Hessian of a logistic regression trained on MNIST. Hessian computed batches of size 64 of the training and test sets, and over the whole train and test sets.}
\label{fig:log_mnist_spacings_ratio}
\end{figure}

\begin{figure}[h]
\centering
\begin{tabular}{cc}
\subfloat[Batch train]{\includegraphics[width=0.4\textwidth]{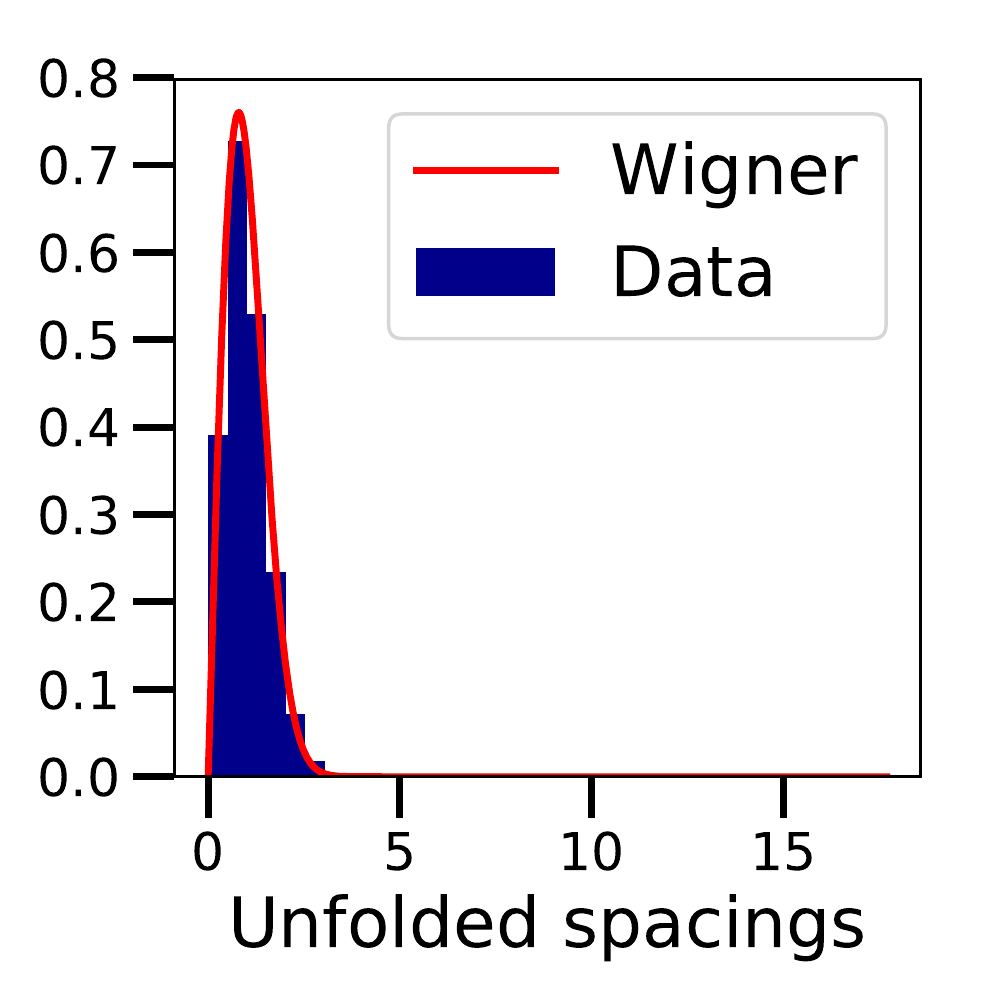}}
& \subfloat[Batch test]{\includegraphics[width=0.4\textwidth]{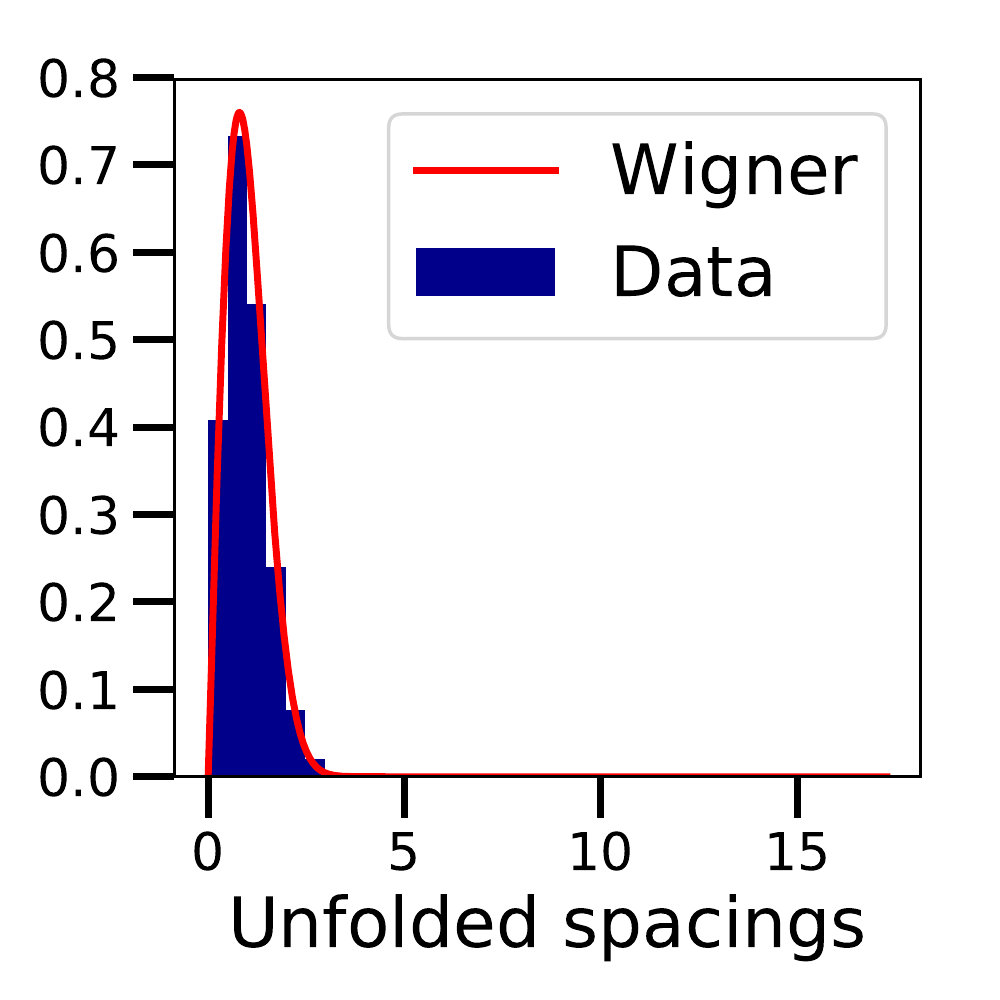}}
\end{tabular}
    \caption{Unfolded spacings for the Hessian of a randomly initialised logistic regression for MNIST. Hessian computed batches of size 64 of the training and test datasets.}
    \label{fig:log_mnist_unt_spacings_unfolded}
\end{figure}

\begin{figure}[h]
\centering
\begin{tabular}{cc}
\subfloat[Batch train]{\includegraphics[width=0.4\textwidth]{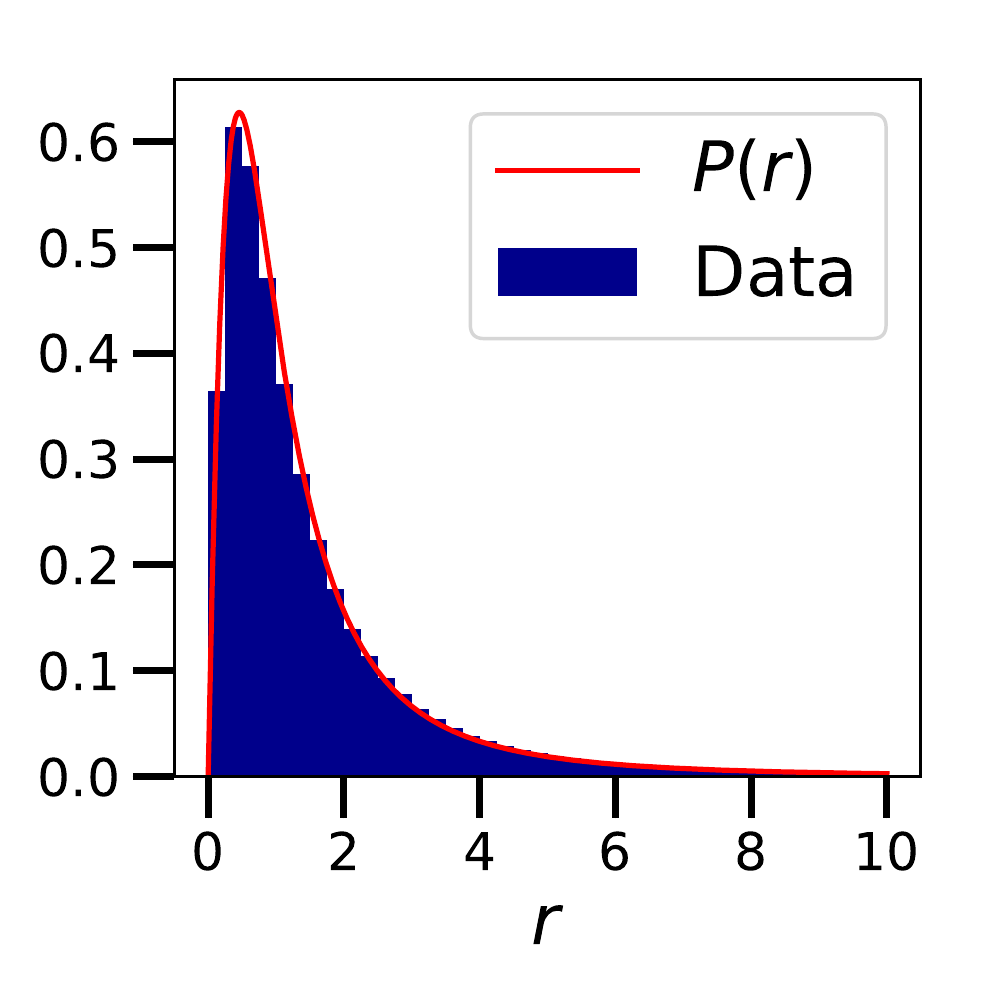}}
& \subfloat[Batch test]{\includegraphics[width=0.4\textwidth]{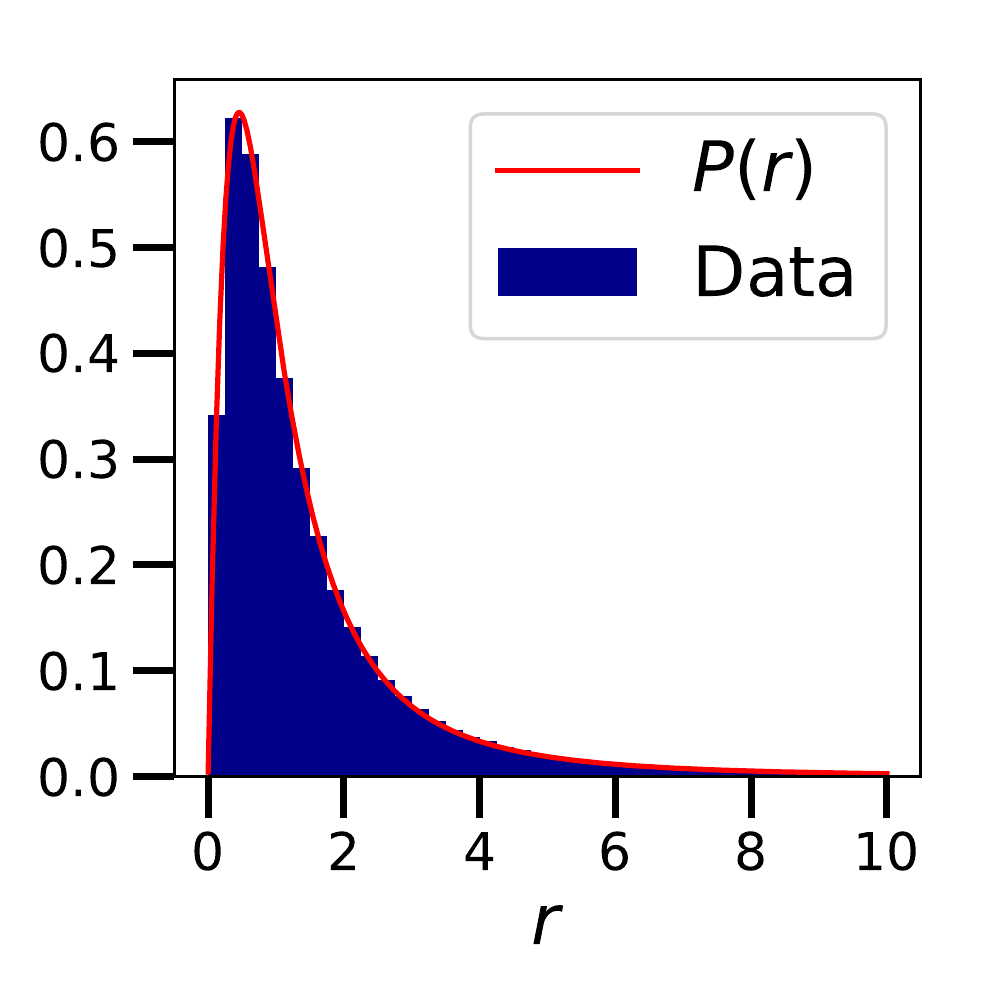}}\\
\subfloat[All train]{\includegraphics[width=0.4\textwidth]{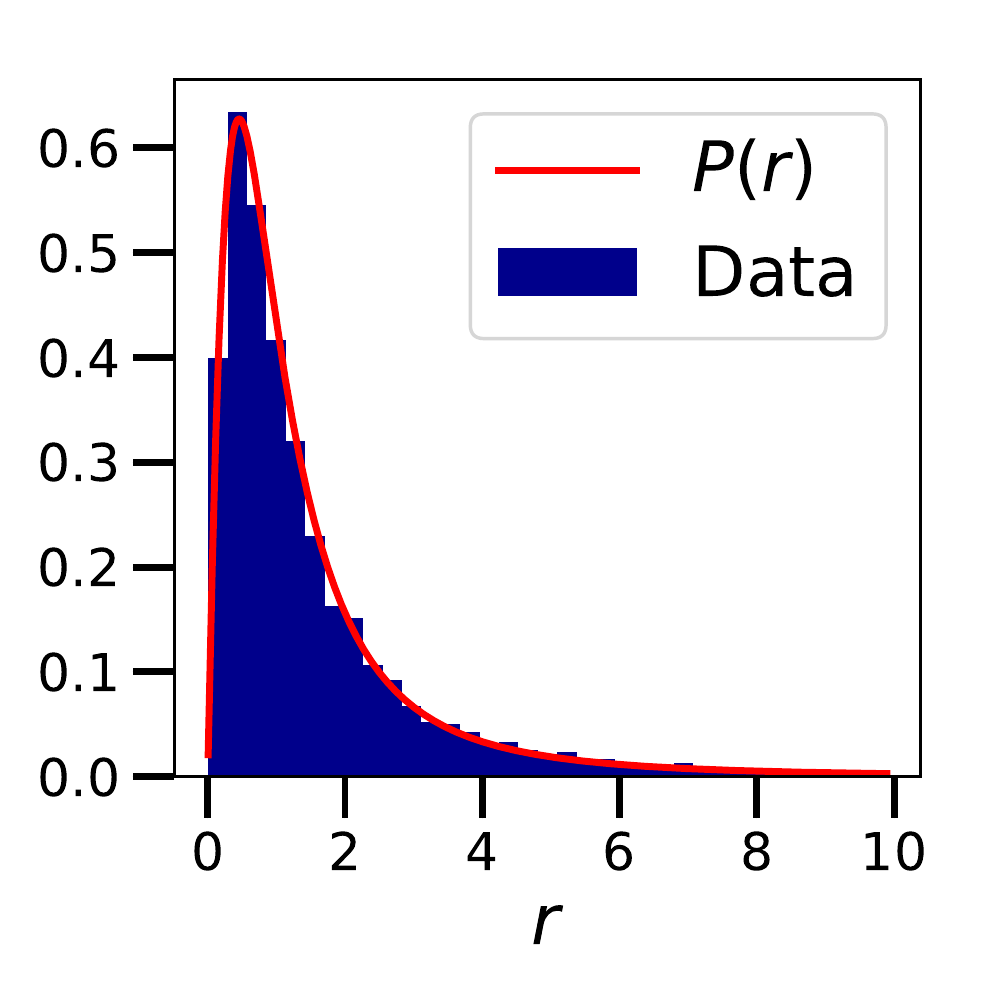}}
& \subfloat[All test]{\includegraphics[width=0.4\textwidth]{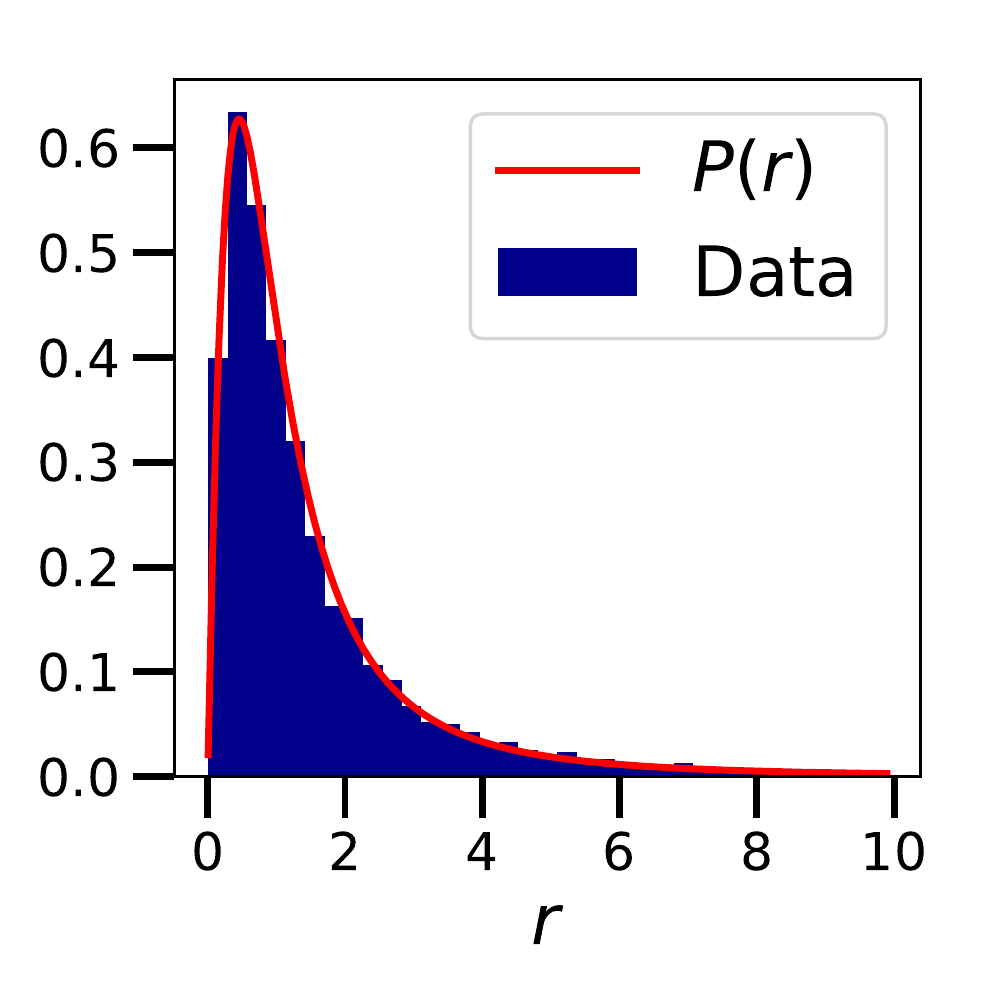}}
\end{tabular}
    \caption{Consecutive spacing ratios for the Hessian of a randomly initialised logistic regression for MNIST. Hessian computed batches of size 64 of the training and test sets, and over the whole train and test sets.}
    \label{fig:log_mnist_unt_spacings_ratio}
\end{figure}

\begin{figure*}[h]
\centering
\begin{tabular}{ccc}
\subfloat[Proportion of small eigenvalues]{\includegraphics[width=0.32\textwidth]{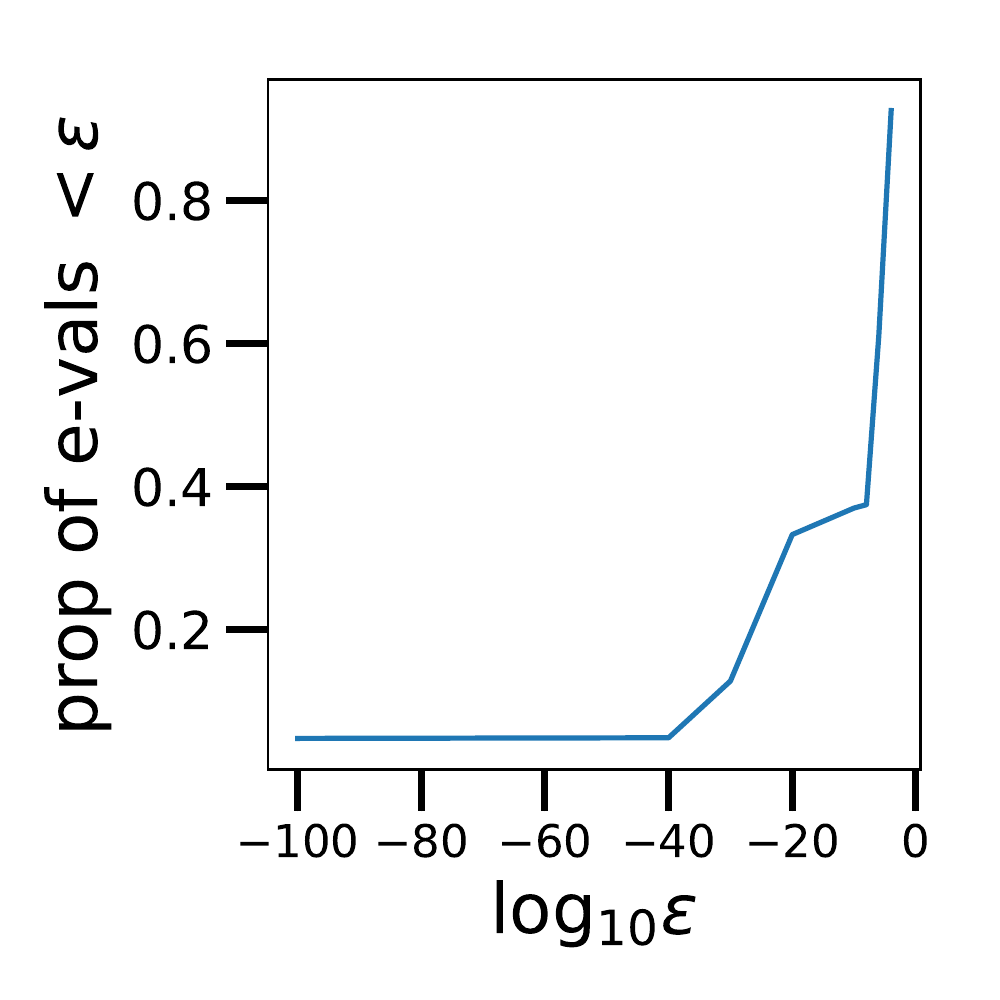}}
& \subfloat[No cut-off]{\includegraphics[width=0.32\textwidth]{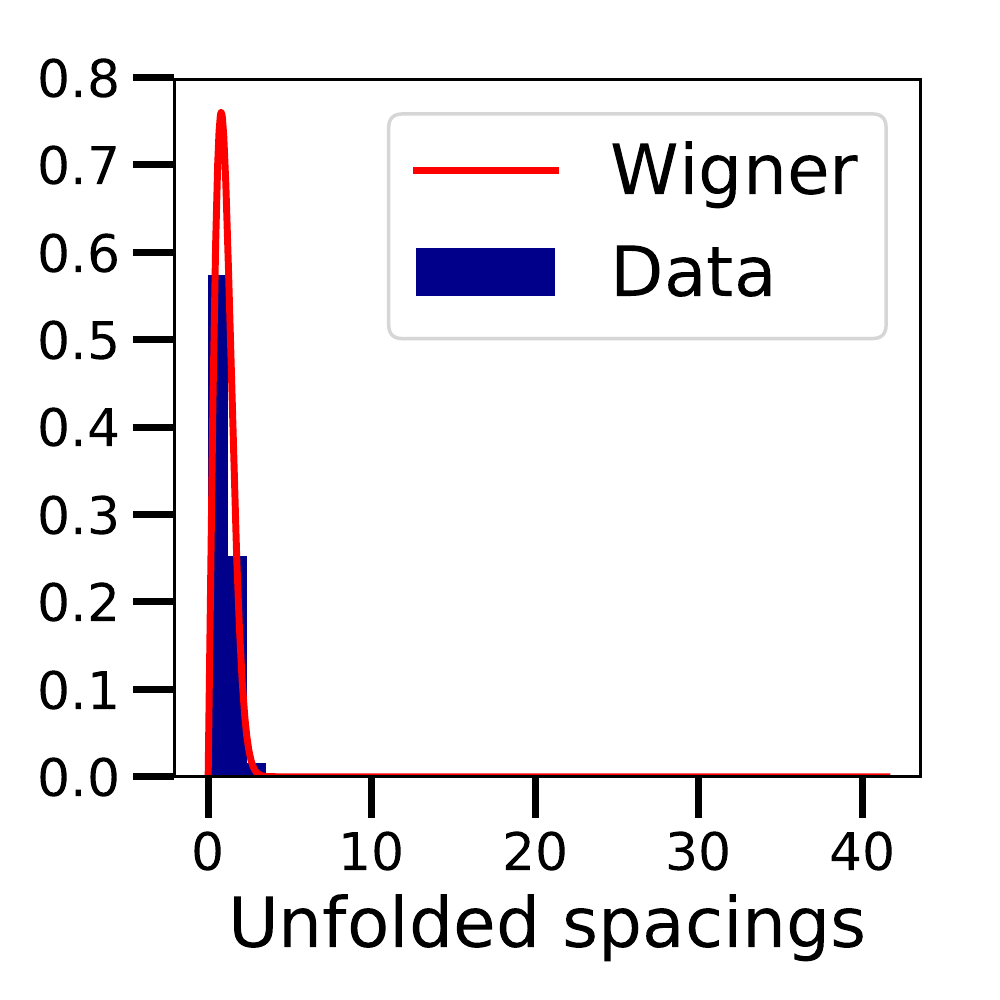}}
& \subfloat[$1e-30$ cut-off]{\includegraphics[width=0.32\textwidth]{3839101829837659314.pdf}}\\

\subfloat[Proportion of small eigenvalues]{\includegraphics[width=0.32\textwidth]{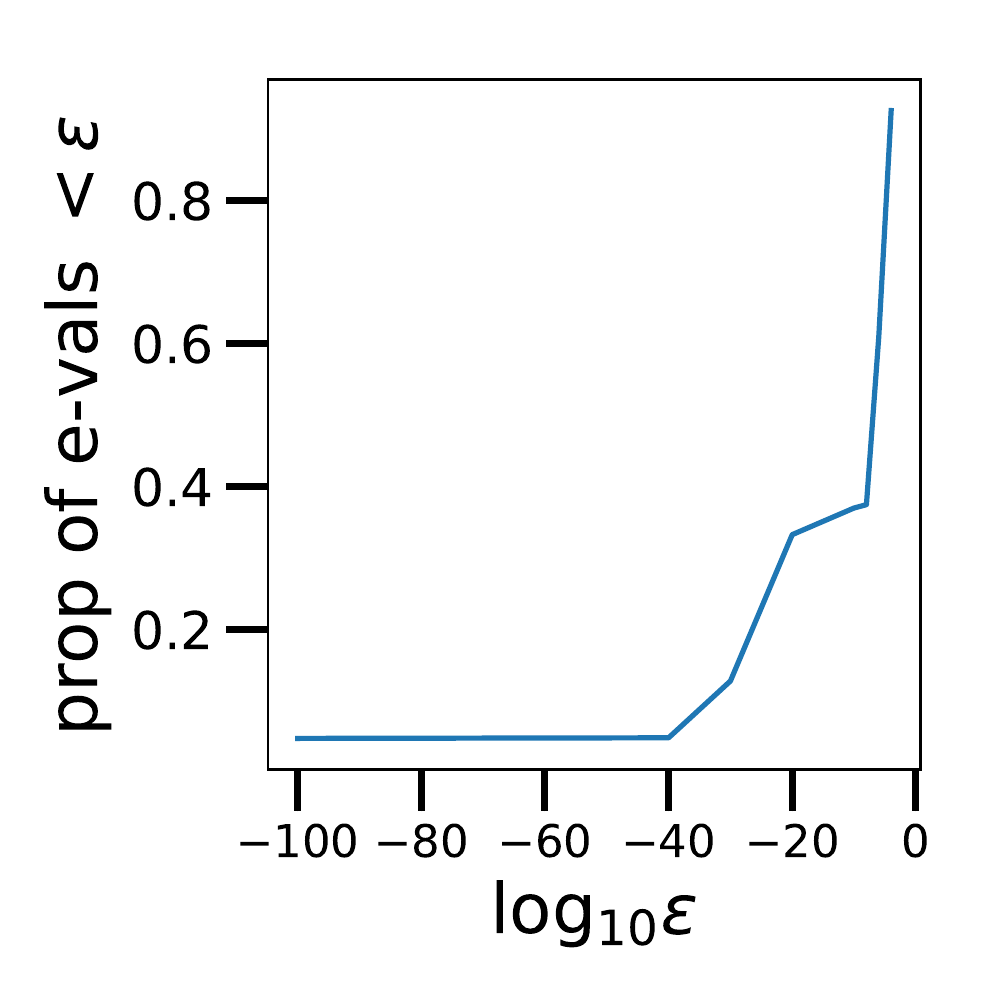}}
& \subfloat[No cut-off]{\includegraphics[width=0.32\textwidth]{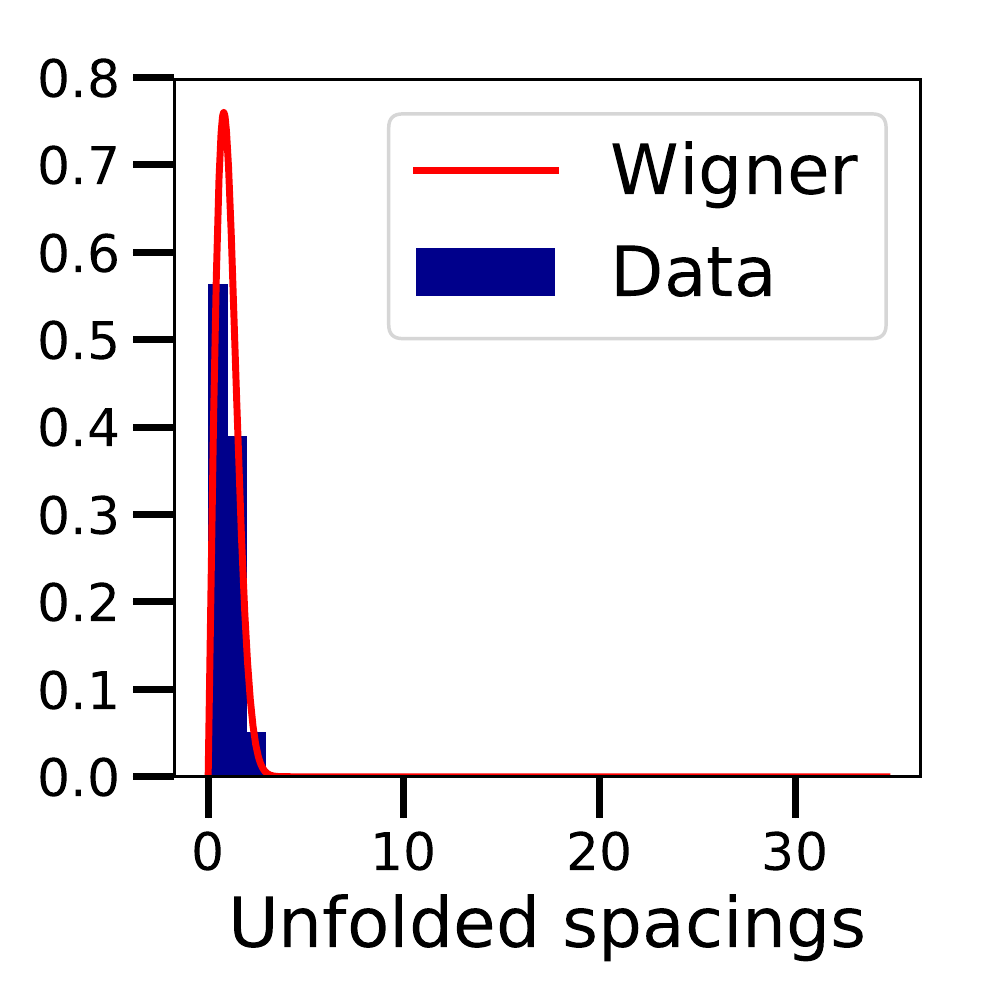}}
& \subfloat[$1e-30$ cut-off]{\includegraphics[width=0.32\textwidth]{_7598967942341315124.pdf}}\\

\subfloat[Proportion of small eigenvalues]{\includegraphics[width=0.32\textwidth]{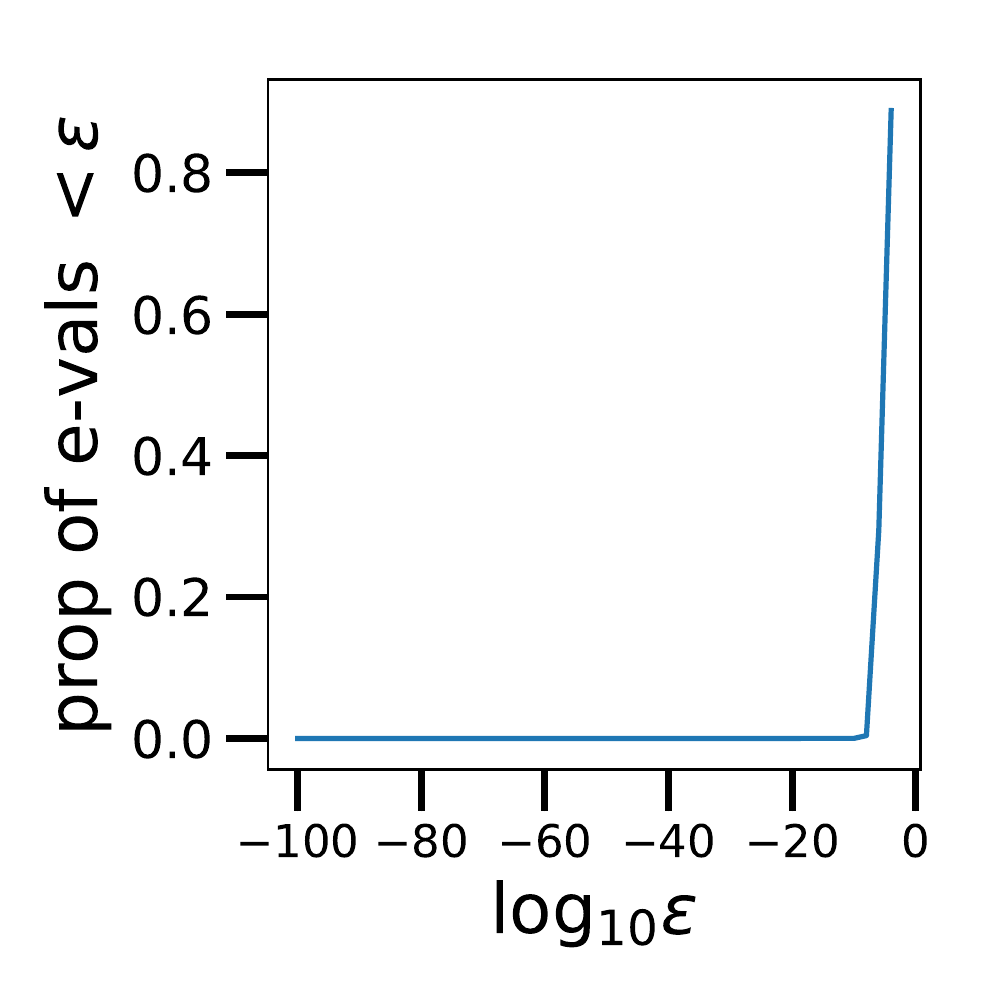}}
& \subfloat[No cut-off]{\includegraphics[width=0.32\textwidth]{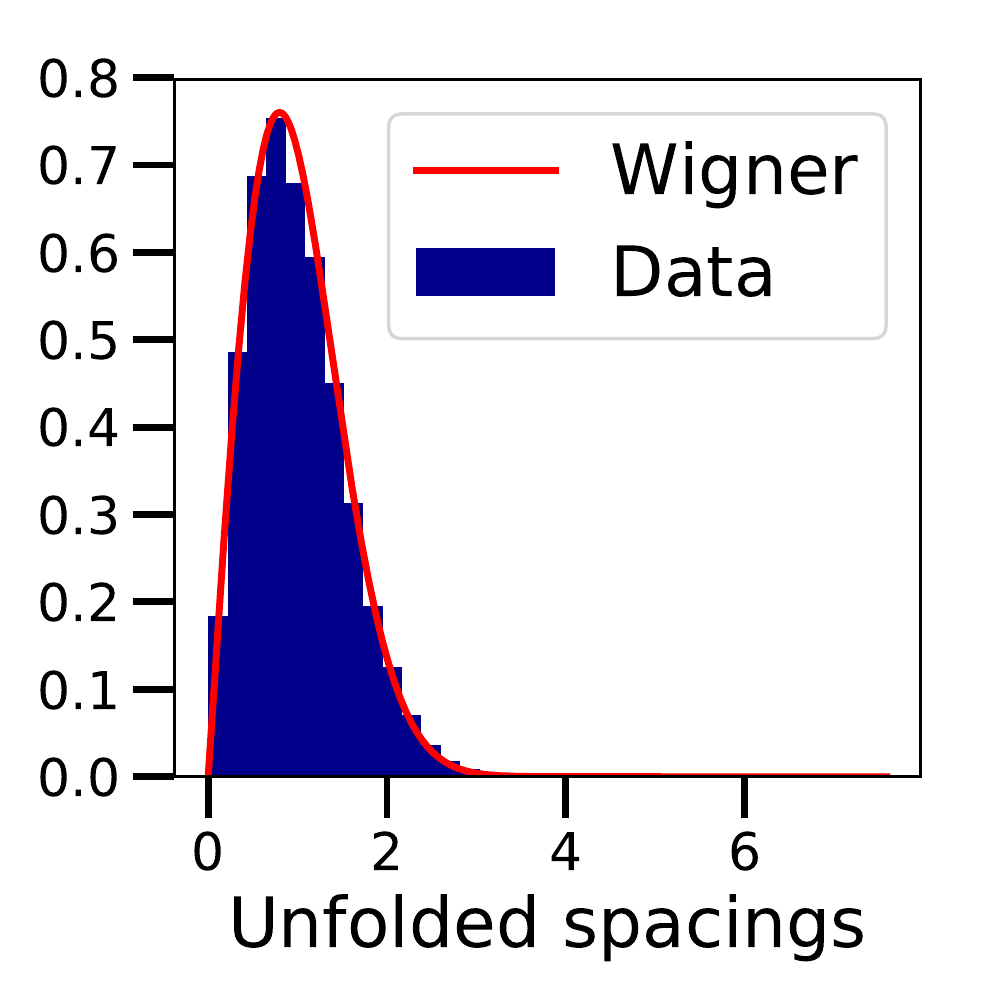}}
& \subfloat[$1e-30$ cut-off]{\includegraphics[width=0.32\textwidth]{_2338619259365618996.pdf}}\\
\end{tabular}
    \caption{Unfolded spacings for the Hessian of a logistic regression. Showing MNIST (top), untrained MNIST (middle) and Resnet34 embedded CIFAR10 (bottom). Comparing the effect of a cuff-off for very small eigenvalues.}
    \label{fig:cifarresnet_comp_mnist_truncation}
\end{figure*}

\bibliographystyle{plainnat}
\bibliography{references}

\section{Experimental details}\label{app:architectures}
\subsection{Network architectures}
\medskip 

\textbf{Logistic regression (MNIST)}
\begin{enumerate}
    \item Input features 784 to 10 output logits.
\end{enumerate}

\medskip

\textbf{2-layer MLP (MNIST)}
\begin{enumerate}
    \item Input features 784 to 10 neurons.
    \item 10 neurons to 100 neurons.
    \item 100 neurons to 10 output logits.
\end{enumerate}

\medskip
\textbf{3-layer MLP (MNIST)}
\begin{enumerate}
    \item Input features 784 to 10 neurons.
    \item 10 neurons to 100 neurons.
    \item 100 neurons to 100 neurons.
    \item 100 neurons to 10 output logits.
\end{enumerate}

\medskip
\textbf{Logistic regression on ResNet features (CIFAR10)}
\begin{enumerate}
    \item Input features 513 to 10 neurons.
\end{enumerate}

\medskip
\textbf{LeNet (CIFAR10)}
\begin{enumerate}
    \item Input features 32x32x3 through 5x5 convolution to 6 output channels.
    \item 2x2 max pooling of stride 2.
    \item 5x5 convolution to 16 output channels.
    \item 2x2 max pooling of stride 2.
    \item Fully connection layer from 400 to 120.
    \item Fully connection layer from 120 to 84.
    \item Fully connection layer from 84 to output 10 logits.
\end{enumerate}

\medskip
\textbf{MLP (CIFAR10)}
\begin{enumerate}
    \item 3072 input features to 10 neurons.
    \item 10 neurons to 300 neurons.
    \item 300 neurons to 100 neurons.
\end{enumerate}

\medskip
\textbf{MLP (Bike)}
\begin{enumerate}
    \item 13 input features to 100 neurons.
    \item 100 neurons to 100 neurons.
    \item 100 neurons to 50 neurons.
    \item 50 neurons to 1 regression output.
\end{enumerate}

\subsection{Other details}
All networks use the same (default) initialisation of weights in PyTorch, which is the `Kaiming uniform' method of \cite{he2015delving}. All networks used ReLU activation functions.

\subsection{Data pre-processing}\label{app:preproc}
For the image datasets MNIST and CIFAR10 we use standard computer vision pre-processing, namely mean and variance standardisation across channels. We refer to the accompanying code for the precise procedure

\medskip
The Bike dataset has 17 variables in total, namely: \texttt{instant}, \texttt{dteday}, \texttt{season}, \texttt{yr}, \texttt{mnth}, \texttt{hr}, \texttt{holiday}, \texttt{weekday}, \texttt{workingday}, \texttt{weathersit}, \texttt{temp}, \texttt{atemp}, \texttt{hum}, \texttt{windspeed}, \texttt{casual}, \texttt{registered}, \texttt{cnt}. All variables are either positive integers or real numbers. It is standard to view \texttt{cnt} as the regressand, so one uses some or all of the remaining features to predict \texttt{cnt}. This is the approach we take, however we slightly reduce the number of features by dropping \texttt{instant}, \texttt{casual}, \texttt{registered}, since \texttt{instant} is just an index and \texttt{casual}+\texttt{registered}=\texttt{cnt}, so including those features would render the problem trivial. We map \texttt{dteday} to a integer uniquely representing the date and we standardise \texttt{cnt} by dividing by its mean.

\end{document}